\documentclass[11pt, journal, web]{article}
\usepackage[margin=16mm]{geometry}
\usepackage{setspace}
\usepackage{times}          
\usepackage[utf8]{inputenc} 

\usepackage{amsmath, amssymb, amsfonts}
\usepackage{mathrsfs}
\usepackage{amsthm}
\usepackage{bbm}            
\usepackage{algorithmic}    

\usepackage{graphicx}       
\usepackage{subcaption}     
\usepackage{caption}        
\usepackage{booktabs}       
\usepackage{array}          
\usepackage{tabularx}       
\usepackage{makecell}       
\usepackage{multirow}       
\usepackage{threeparttable} 
\usepackage{wrapfig}        

\usepackage[ruled,vlined]{algorithm2e}

\usepackage{hyperref}
\hypersetup{
   colorlinks=true,
   linkcolor=blue,
   citecolor=blue,
   urlcolor=blue
}
\usepackage{cleveref}       

\usepackage[normalem]{ulem} 
\useunder{\uline}{\ul}{}    
\usepackage{url}            
\usepackage{verbatim}       

\usepackage{pifont}         
\usepackage{utfsym}         

\usepackage{rotating}       
\usepackage{pdflscape}      

\usepackage{authblk}        

\crefformat{figure}{\color{blue}#2Fig.~#1#3}

\makeatletter
\renewcommand\AB@affilsepx{, \protect\Affilfont}
\makeatother
\usepackage{tabularx}
\providecommand{\keywords}[1]
{
  \small	
  \textbf{\textit{Keywords---}} #1
}

\begin{document}

\title{\textbf{Enhancing Semi-Supervised Multi-View Graph Convolutional Networks via Supervised Contrastive Learning and Self-Training }}
\author[1]{Huaiyuan Xiao}
\author[1, 2]{Fadi Dornaika \thanks{Corresponding author}}
\author[3]{Jingjun Bi}
\affil[1]{\textit{University of the Basque Country}}
\affil[2]{\textit{IKERBASQUE}}
\affil[3]{\textit{North China University of Water Resources and Electric Power}}

\affil[ ]{

\small\texttt{huaiyuan.xiao@foxmail.com, fadi.dornaika@ehu.eus, bi.jingjun@outlook.com}}
\date{}
\maketitle

\date{}
\maketitle

\begin{abstract}
The advent of graph convolutional network (GCN)-based multi-view learning provides a powerful framework for integrating structural information from heterogeneous views, enabling effective modeling of complex multi-view data. However, existing methods often fail to fully exploit the complementary information across views, leading to suboptimal feature representations and limited performance. To address this, we propose MV-SupGCN, a semi-supervised GCN model that integrates several complementary components with clear motivations and mutual reinforcement. First, to better capture discriminative features and improve model generalization, we design a joint loss function that combines Cross-Entropy loss with Supervised Contrastive loss, encouraging the model to simultaneously minimize intra-class variance and maximize inter-class separability in the latent space. Second, recognizing the instability and incompleteness of single graph construction methods, we combine both KNN-based and semi-supervised graph construction approaches on each view, thereby enhancing the robustness of the data structure representation and reducing generalization error. Third, to effectively utilize abundant unlabeled data and enhance semantic alignment across multiple views, we propose a unified framework that integrates contrastive learning in order to enforce consistency among multi-view embeddings and capture meaningful inter-view relationships, together with pseudo-labeling, which provides additional supervision applied to both the cross-entropy and contrastive loss functions to enhance model generalization. Extensive experiments demonstrate that MV-SupGCN consistently surpasses state-of-the-art methods across multiple benchmarks, validating the effectiveness of our integrated approach. The source code is available at \url{https://github.com/HuaiyuanXiao/MVSupGCN}
\end{abstract}

 \hspace{10pt}

\section{Introduction}

Multi-view data consists of information collected from different perspectives or feature sets, providing a fuller and more accurate depiction of the same object. Despite originating from different views, the data inherently shares the same semantic information while maintaining complementarity among views, enabling the extraction of unique and relevant features \cite{Chen_2023}. 

In practice, multi-view data has been widely applied in various real-world domains, including data mining \cite{2020Efficient,WU2025111764,2019Adaptive}, machine learning \cite{DORNAIKA2025107218,2021Multi,xu2024reliable}, and computer vision \cite{10658393,LIN2024111920,2020cmSalGAN}. While individual views offer only partial representations, their integration conveys a more complete and consistent understanding of the underlying entity. This inherent consistency and complementarity often result in superior performance compared to single-view data, serving as a key motivation for the growing research interest in multi-view learning \cite{2021Differentiable,2020CGD,2021Generalized}.

Collecting labeled data for multi-view datasets is both time-intensive and resource-demanding. Despite their richness in information, multi-view datasets often lack sufficient labeled samples. This limitation highlights the need to harness unlabeled data to address the shortage of annotations. Studies have shown that leveraging unlabeled multi-view data can yield excellent results \cite{hu2024deep}, positioning multi-view semi-supervised learning as a critical research focus.

In the last decade, multi-view learning has emerged as a prominent research direction, demonstrating its effectiveness across supervised, semi-supervised, and unsupervised learning scenarios \cite{he2023similarity}. Numerous methodological frameworks have been introduced to address the challenges posed by multi-view data. Among these, Co-training \cite{1998Combining,2010A,kumar2011co} and Co-regularization \cite{2008An,2011Active} are widely recognized for leveraging complementary information from different views. Multiple Kernel Learning approaches \cite{liu2023contrastive} have been proposed to integrate heterogeneous feature representations, while subspace learning methods \cite{white2012convex,li2019reciprocal} aim to discover shared latent structures among views.

Despite their contributions to multi-view learning, the majority of these methods are still confined to shallow architectural designs. Consequently, there is growing interest in leveraging the greater representational capacity of deep learning to achieve more robust and satisfactory outcomes. Deep learning models are particularly effective at processing complex data and recognizing patterns\cite{wu2020deep,yan2021deep}, as they can extract sophisticated features through multiple layers of non-linear transformations. Among the numerous deep learning techniques, graph-based approaches stand out for their unique advantage of representing arbitrarily distributed data. In general, graph-based multi-view learning approaches utilize graphs that represent the pairwise similarities among samples, serving as the basis for the development of numerous related methods.

Graph Convolutional Networks (GCNs)\cite{kipf2016semi}, originally developed for single-view data, have shown remarkable effectiveness in tasks like node and graph classification. Integrating multi-view learning with GCNs offers a promising approach to comprehensively modeling complex data. Within a multi-view \textcolor{black}{Graph Convolutional Network} framework, each view provides a distinct graph topology and corresponding node attributes, which are processed through specialized GCN branches. A fundamental component of this methodology is the efficient integration of the heterogeneous feature representations produced by all branches. By employing advanced fusion strategies to aggregate these diverse embeddings into a unified model, the framework achieves a more holistic and in-depth interpretation of the underlying data. This design allows multi-view GCNs to simultaneously extract view-specific features and discover commonalities among views, thereby supporting efficient learning even when only a small amount of labeled data is available.

Currently, numerous researchers have explored the use of \textcolor{black}{Graph Convolutional Networks} for semi-supervised learning on multi-view graphs \cite{2020Co,wu2023interpretable,chen2023learnable, PENG2025127194}. However, there are still several research gaps, particularly in the following areas:

1. In multi-view classification tasks, the large number of classes combined with the limited number of samples within each class often poses challenges for deep learning methods, resulting in certain performance limitations. \textcolor{black}{Specifically, GCN-based approaches suffer from insufficient intra-class feature compactness and weak inter-class separability due to inadequate constraints during training, which leads to suboptimal classifier optimization and degraded generalization.}

2. Despite numerous studies in the field of graph contrastive learning, multi-view graph contrastive learning, particularly at the level of graph construction, remains underexplored and relatively lacking in existing research. \textcolor{black}{Most existing works apply contrastive learning on single-view or node-level embeddings, ignoring the opportunity to exploit complementary structural information from multiple graph construction strategies, thus limiting representation richness.}

3 \textcolor{black}{In semi-supervised learning, although pseudo-labeling provides partial supervisory signals for unlabeled data, existing methods typically make limited use of unlabeled samples, relying mainly on pseudo-label supervision without effectively exploring the underlying structural and semantic information within the unlabeled data. As a result, the potential of unlabeled samples is not fully exploited, restricting the improvement of model accuracy and generalization.}

To address the challenges of multi-view learning, this paper introduces a novel framework that efficiently integrates information from multiple graph views named Multi-View Graph Convolutional Network Enhanced with Supervised Contrastive Loss for Semi-Supervised Learning (MV-SupGCN). The proposed Mv-SupGCN method enhances model generalization by integrating multiple innovative techniques. It combines a joint loss function with both Cross-Entropy and SupCon losses to measure intra- and inter-class distances. \textcolor{black}{The method employs two distinct graph construction approaches—K-Nearest Neighbors (KNN) and a semi-supervised graph—on the same view to capture complementary structural information, thereby enriching node relationships and stabilizing classifier learning through increased representation diversity.} Additionally, by applying contrastive learning and pseudo-labeling to unlabeled data, Mv-SupGCN maximizes the potential of unlabeled samples, further boosting the model's accuracy and effectiveness.  

The main contributions are as follows:

1. \textcolor{black}{We propose a joint loss function combining Cross-Entropy and Supervised Contrastive losses that effectively enhances intra-class compactness and inter-class discrimination, addressing the key issue of insufficient class-wise constraints in existing GCN-based methods.}

2 \textcolor{black}{Two distinct graph construction methods—K-Nearest Neighbors and a semi-supervised graph approach—are applied to the same view to capture complementary structural information. Specifically, KNN effectively captures local neighborhood relationships, emphasizing fine-grained, instance-level connections, while the semi-supervised graph leverages label information to reveal higher-order, global semantic structures. By integrating these diverse graph construction strategies, the model enriches the representation space with both local and global insights, increasing classifier diversity and contributing to a more robust and stable hypothesis space. This innovative multi-graph fusion enables more comprehensive modeling of complex data relationships than traditional single-graph approaches.}

3. \textcolor{black}{A novel end-to-end framework for semi-supervised graph representation learning is proposed, which integrates pseudo-labeling with both contrastive learning and cross-entropy loss to fully exploit unlabeled data. By generating pseudo-labels, the method provides supervisory signals for unlabeled samples and incorporates them alongside labeled data in both the contrastive learning and cross-entropy optimization processes. Furthermore, contrastive learning is innovatively applied to multiple views of unlabeled samples, establishing meaningful connections between these views and effectively capturing their underlying semantic structure. This dual strategy not only boosts classification accuracy and generalization ability, but also overcomes the limitations of existing contrastive learning approaches that neglect multi-view structural diversity.}

4. To assess the effectiveness of the proposed method, we conducted a series of experiments using seven publicly available multi-view datasets and compared its performance with leading graph-based algorithms. The findings demonstrate that our approach consistently outperforms existing methods.

The remainder of this paper is organized as follows. Section \ref{sec:relatedwork} reviews recent advancements in relevant research domains. Section \ref{sec:background} introduces fundamental concepts and background knowledge essential to this study. In Section \ref{sec:proposedmodel}, we provide an in-depth exposition of the proposed MV-SupGCN approach. Section \ref{sec:experimentalsetup} outlines the experimental design and implementation details. The experimental findings are analyzed and discussed in Section \ref{sec:experimentalresults}. Lastly, Section \ref{sec:conclusion} provides a summary of the paper's main contributions, discussing possible directions for future work and applications, and considering the limitations of the current approach.

\section{Related Work}
\label{sec:relatedwork}
\normalsize\textit{\subsection{Multi-view Learning}}

Multi-view learning seeks to enhance model performance by efficiently integrating data from various sources or different representations. This area has fostered the development of numerous seminal algorithms, each exhibiting robust effectiveness across a wide range of practical applications. Co-training \cite{1998Combining} is an early multi-view learning approach that improves classifiers by using two independent views and unlabeled data. Multiple Kernel Learning (MKL) \cite{G2011Multiple} enhances results by fusing information from different views through optimized kernel functions. \textcolor{black}{Multi-view Clustering (MVC) \cite{10108535} integrates complementary information from multiple perspectives to effectively group data samples.}

As deep learning has progressed, Multi-view Graph Convolutional Networks have become a leading approach in multi-view learning. These networks capture local structure within each view through graph convolutions, while also modeling global patterns by analyzing the relationships among multiple views. For instance, Co-GCN \cite{2020Co} combines co-training with graph-based learning, leveraging a fused Laplacian matrix from multiple KNN-based views to enhance semi-supervised learning. WFGSC \cite{bi2024sample} constructs a semi-supervised graph for each view, jointly estimating graphs and labels, and fuses them into a unified graph to aggregate multi-view information.  LGCNFF \cite{chen2023learnable} introduces an end-to-end framework integrating sparse autoencoders with an adaptive GCN for enhanced feature extraction and classification accuracy. Similarly, the MGCN-FN\cite{Peng2025} framework is designed to efficiently integrate multi-view feature information and graph structures, thereby improving the performance of semi-supervised multi-view classification tasks. GEGCN \cite{10462517} is a generative essential graph convolutional network that integrates multi-graph consistency and complementarity extraction, graph refinement, and classification into a unified optimization framework for multi-view semi-supervised classification. Lastly, MV-TriGCN \cite{XIAO2025103420} integrates enhanced triplet loss-based deep metric learning into a graph convolutional framework, enabling robust and accurate representation across diverse data views.

\normalsize\textit{\subsection{Contrastive Learning}}

Contrastive learning has significantly advanced the field of self-supervised representation learning. These approaches primarily depend on comparing many pairs of representations. Their objective is to enhance the similarity between positive pairs while decreasing it for negative pairs within a latent feature space. In the context of multi-view learning, positive pairs consist of invariant representations derived from multiple views of the same sample, whereas negative pairs come from invariant representations of different samples across various views.

SupCon \cite{khosla2020supervised} leverages class label information within a contrastive learning framework to improve representation learning in supervised scenarios. This enhancement broadens the applicability of contrastive learning, making it effective in both supervised and semi-supervised settings. 

TGNN \cite{ijcai2022p295} employs two separate graph neural networks to capture diverse perspectives when learning graph representations.
InfoGraph \cite{c921e40702284b7892beb56a62ed32b5} constructs node representations through message passing and then aggregates them to produce a comprehensive graph-level representation.
\textcolor{black}{Bi-CLKT \cite{SONG2022108274} is a knowledge tracing algorithm that uses bi-graph contrastive learning to jointly model exercises and concepts, achieving more accurate predictions of student knowledge mastery.}

In the domain of multi-view clustering, contrastive learning plays a crucial role in managing feature alignment and representation learning across diverse views. CVCL \cite{Chen_2023} introduces a cross-view contrastive learning approach that focuses on learning view-invariant representations and derives clustering results by contrasting cluster assignments from multiple views. Furthermore, MVC-PGCL-DCL \cite{hu2024deep} proposes an innovative deep multi-view clustering framework that integrates pseudo-label guided contrastive learning with dual correlation learning to enhance clustering performance.

\textcolor{black}{Building upon these prior works, our approach distinctly differs by simultaneously leveraging both KNN and semi-supervised graphs to capture complementary local and global structural information within a unified semi-supervised multi-view GCN contrastive learning framework.  Unlike existing methods that focus on single-view data or lack explicit graph-level representation learning, our framework integrates multi-graph fusion with joint supervised contrastive and classification losses using the origianl labels and the adaptive pseudo-labels, thereby enhancing representation discriminability and effectively exploiting unlabeled data. This combination addresses the limitations of previous approaches and enables more robust and comprehensive multi-view graph representation learning.}

\section{Background }
\label{sec:background}
\normalsize\textit{\subsection{Notations}}

This section begins by introducing the key mathematical notations used throughout the manuscript, followed by a discussion of models relevant to our study. We use bold lowercase letters to denote vectors and bold uppercase letters to denote matrices in this paper.

In a single-view setting, we denote the data matrix as $\mathbf{X} = [\mathbf{x}_{1}; \mathbf{x}_{2}; \ldots; \mathbf{x}_{n}] \in \mathbb{R}^{n \times d}$, where each of the $n$ rows corresponds to a sample with $d$ features. The graph adjacency matrix $\mathbf{A} \in \mathbb{R}^{n \times n}$ encodes the pairwise similarities among the samples in $\mathbf{X}$. The degree matrix $\mathbf{D}$ is a diagonal matrix where each entry is the sum of the corresponding row in $\mathbf{A}$. The identity matrix is denoted by $\mathbf{I}$. The normalized Laplacian matrix is then given by $\mathbf{L} = \mathbf{I} - \mathbf{D}^{-\frac{1}{2}} \mathbf{A} \mathbf{D}^{-\frac{1}{2}}$.

For a multi-view dataset, there are $V$ different views, and each one contains $n$ samples. In each view, a sample is encoded as a row vector with $d_v$ features. The data matrix for the $v$-th view is expressed as $\mathbf{X}_v = [\mathbf{x}_{v1}; \mathbf{x}_{v2}; \ldots; \mathbf{x}_{vn}] \in \mathbb{R}^{n \times d_v}$.

The ground truth labels are  represented by the matrix $\mathbf{Y} \in \mathbb{R}^{n \times c}$, where each column corresponds to one of the $c$ classes. In the Graph Convolutional Network, each layer $l$ has a trainable weight matrix $\mathbf{W}^{(l)} \in \mathbb{R}^{d_l \times d_{l+1}}$. The initial feature dimension is given by $d_0 = d$. The GCN produces a prediction matrix $\mathbf{Z} \in \mathbb{R}^{n \times c}$, where each entry represents the predicted probability of a sample belonging to a particular class.

\normalsize\textit{\subsection{Graph Convolution Network}}

Consider an undirected graph $\mathcal{G}=\{\mathbf{X},\mathbf{A}\}$, where $\mathbf{A}$ is the adjacency matrix. It can be constructed using the KNN algorithm. The adjacency matrix $\mathbf{A} \in \mathbb{R}^{n \times n}$ represents the pairwise connections among samples in the dataset, which is described by the feature matrix $\mathbf{X}$. The propagation mechanism of node embeddings over $L$ layers in a GCN is formally defined as:
\begin{equation}
    \mathbf{H}^{(l+1)}=\sigma(\mathbf{D}^{-\frac12}\widetilde{\mathbf{A}}\mathbf{D}^{-\frac12}\mathbf{H}^{(l)}\mathbf{W}^{(l)})
\label{eq:H(l+1)}
\end{equation}

Let $\widetilde{\mathbf{A}}=\mathbf{A}+\mathbf{I}$ denote the adjacency matrix augmented with self-connections, where $\mathbf{I}$ represents the identity matrix. The degree matrix of $\widetilde{\mathbf{A}}$   is represented as $\mathbf{D}$. The activation function is denoted as $\sigma(\cdot)$ which can be a non-linear function such as sigmoid or ReLU. The output of the $l$-th layer is denoted by $\mathbf{H}^{(l+1)}\in\mathbb{R}^{n\times d_{(l+1)}}$,  where the input to the network is defined as $\mathbf{H}^0=\mathbf{X}$.
The output of the final layer in the Graph Convolutional Network is formally expressed as: 
\begin{equation}
\mathbf{Z}=softmax(\mathbf{D}^{-\frac{1}{2}} \mathbf{ \widetilde{A}}  \,  \mathbf{D}^{-\frac{1}{2}}\mathbf{H}^{(L-1)}\mathbf{W}^{(L-1)})
\end{equation}

The set of learnable weight matrices, $\{\mathbf{W}^{(0)}, \mathbf{W}^{(1)}, \ldots, \mathbf{W}^{(L-1)}\}$, is optimized by minimizing the cross-entropy loss computed over the labeled nodes. The loss function is defined as:
\begin{equation}
\mathcal{L}_{Semi-GCN}=-\sum_{i\in\Omega}\sum_{j=1}^cY_{ij} \log  Z_{ij}
\end{equation}

where $\Omega$ represents the set of labeled node indices.

\section{Proposed Model}
\label{sec:proposedmodel}

\begin{sidewaysfigure*}
    \centering
    \includegraphics[width=0.95\linewidth]{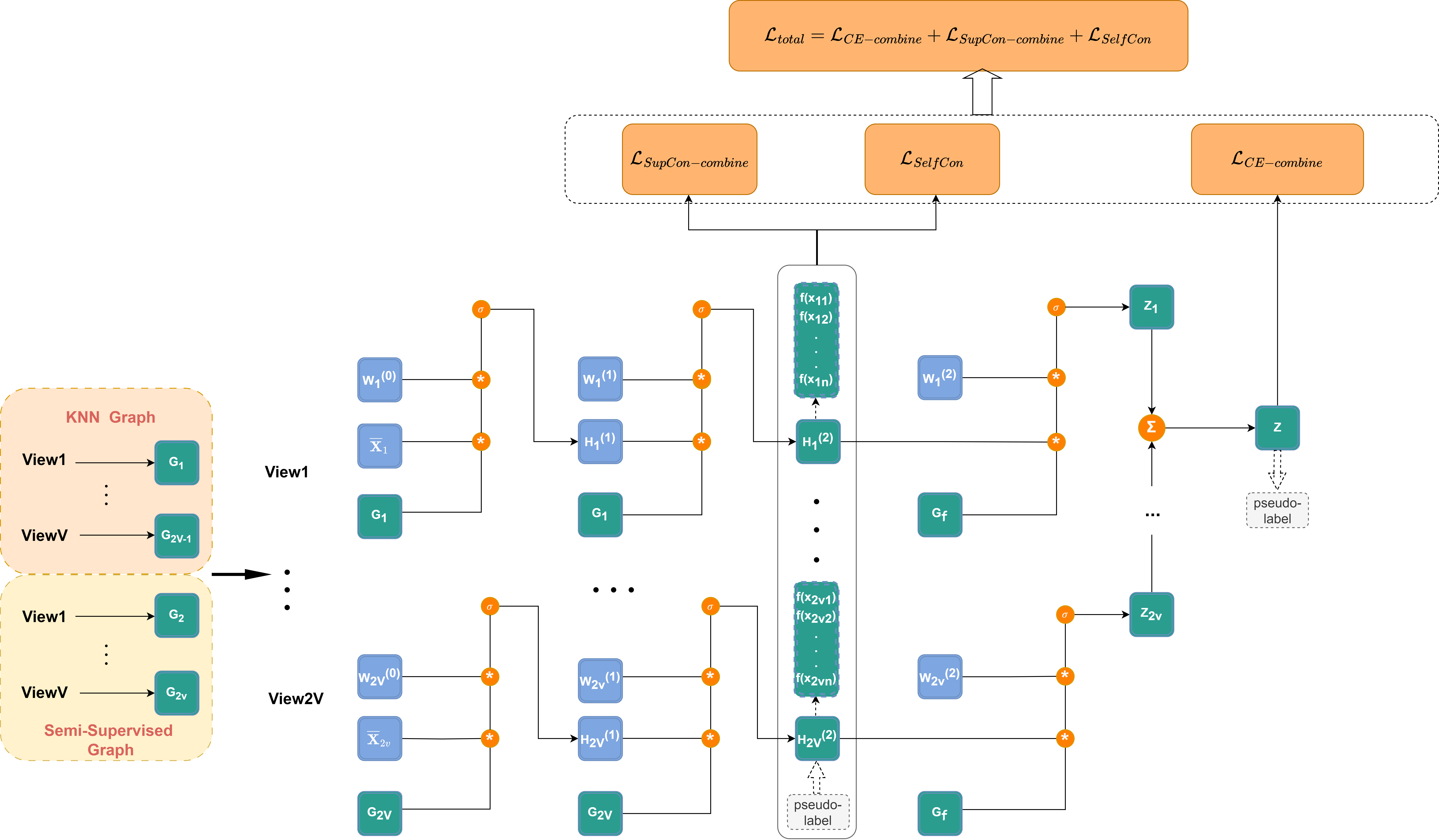}
    \caption{\textcolor{black}{The MV-SupGCN architecture consists of $2V$ GCNs, each assigned to a unique data view and its corresponding graph. Specifically, each view is represented by two distinct graphs. Each GCN within this framework is composed of three layers, enabling the model to perform hierarchical feature extraction and progressively capture complex representations from the input data.}}
    \label{fig: MVSupGCN}
\end{sidewaysfigure*}

Graph Convolutional Networks were originally developed for single-view and face considerable difficulties when applied to multi-view scenarios. Effectively adapting GCNs for multi-view data remains an open challenge in the field. In this section, we present a new method to address this gap. Fig. \ref{fig: MVSupGCN} provides an overview of the proposed model MV-SupGCN.

\normalsize\textit{\subsection{Multi-view GCN Architecture}}

We present a multi-view GCN classifier with two core modules to enhance classification performance. The first component consists of multiple GCN branches, where each view is modeled using two distinct GCNs constructed from different graph structures (e.g., KNN and semi-supervised graphs). Each GCN serves as an independent classifier, enabling diverse representations within the same view. The second component integrates information across all views using an aggregation mechanism. The final prediction is obtained through an ensemble learning strategy that combines the outputs of all view-specific classifiers.

\normalsize\textit{\subsubsection{View-specific Graph Construction}}

For each view, we construct two distinct types of graphs. The first graph is generated using the classical unsupervised K-nearest neighbors algorithm, which establishes connections between each sample and its K most similar counterparts. In contrast, the second graph is constructed using a semi-supervised approach that leverages both labeled and unlabeled data to better capture additional structural information.

\textbf{a. View-based  KNN graphs}

When applying graph-based algorithms to real-world multi-view data, it is crucial to construct appropriate graphs, as such data often lacks an inherent topology. Among various graph construction methods, the KNN algorithm is widely used and has demonstrated promising results. Let $\mathcal{G} = \{\mathbf{X}, \mathbf{A}\}$, where $\mathbf{X}$ represents the feature matrix and $\mathbf{A}$ denotes the adjacency matrix. The element at position $(i, j)$ in $\mathbf{A}$ is defined as follows:
\begin{equation}
\mathbf{A}_{ij}=\begin{cases}1,\mathbf{x}_i\in\mathrm{KNN}\big(\mathbf{x}_j\big) \; \mathrm{or} \; \mathbf{x}_j\in\mathrm{KNN}(\mathbf{x}_i),\\0,\mathrm{otherwise},\end{cases}
\end{equation}
where $\mathbf{x}_i$ denotes the $i$-th row vector of $\mathbf{X}$. KNN($\mathbf{x}_i$) represents the set of $k$ nearest neighbors of $\mathbf{x}_i$, and $\mathbf{A}_v$ refers to the $v$-th adjacency matrix. After the renormalization step, each matrix transforms into:
\begin{equation}
\widehat{\mathbf{A}}_v=\mathbf{D}^{-\frac{1}{2}}(\mathbf{I}+\mathbf{A}_v)\mathbf{D}^{-\frac{1}{2}}
\label{eq:Ahat}
\end{equation}
The flexible graph filter corresponding to the $v$-th view can be computed as follows:
\begin{equation}
\overline{{\mathbf{A}}}_v=(1-\beta)\mathbf{I}+\beta{\widehat{\mathbf{A}}}_{v}
\label{eq:tildeAv}
\end{equation}

\textbf{b. View-based semi-supervised graph $\mathbf{S}_v$}

A semi-supervised approach is employed to construct a graph similarity matrix for each data view, yielding $V$ graphs $\mathbf{S}_v$ that capture pairwise sample similarities within their respective views $\mathbf{X}_v$. This method enables the propagation of labeled information, thereby facilitating the acquisition of more comprehensive global information.

We construct the matrix for each graph $\mathbf{S}_v$ using a modified variant of the shallow semi-supervised framework MVCGL proposed in \cite{ZIRAKI2022174}, which builds a fused graph from all view data. Specifically, the graph $\mathbf{S}_v$ is estimated by optimizing the following objective function:

\begin{equation}
\begin{aligned}&\min_{\mathbf{S}_{v},\mathbf{O},\mathbf{Q}_{v},\mathbf{b}_v}Trace(\mathbf{X}_{v}^{T}\mathbf{L}_{v}\mathbf{X}_{v})+\eta\:Trace(\mathbf{O}^{T}\mathbf{L}_{v}\mathbf{O})\\&+Trace((\mathbf{O}-\mathbf{Y})^{T}\mathbf{U}(\mathbf{O}-\mathbf{Y})\\&+\gamma\left\|\mathbf{S}_{v}\right\|^{2}+\mu(\left\|\mathbf{Q}_{v}\right\|^{2}+\alpha\left\|\mathbf{X}_{v}\mathbf{Q}_{v}+\mathbf{1}\mathbf{b}_v^{T}-\mathbf{O}\right\|^{2})\quad \\&s.t.\quad0\leq S_{v(ij)}\leq1,\sum_{j=1}^{N}S_{v(ij)}=1\end{aligned}
\label{eq:SGraph}
\end{equation}

Here, $\mathbf{O}$ denotes the soft label predictions corresponding to $\mathbf{X}_v$, while $\mathbf{L}_v$ refers to the normalized Laplacian matrix constructed from $\mathbf{S}_v$. The projection matrix, $\mathbf{Q} \in \mathbb{R}^{d_{v}\times c}$, and the bias vector, $\mathbf{b}$, give a linear mapping between feature space and label space.  The diagonal matrix $\mathbf{U}$ assigns non-zero values to labeled samples and zeros elsewhere.  The solution is obtained via simple alternating minimization, similar to the method used in \cite{ZIRAKI2022174}. Our objective is to obtain the graph similarity matrix $\mathbf{S}_v$.

The main advantages of this semi-supervised approach are: (1) the first two terms in the objective function regulate the graph matrix $\mathbf{S}_v$ to ensure both feature and label smoothness; (2) it allows for the simultaneous estimation of both the graph structure and the labels of unlabeled samples; and (3) the linearization term promotes greater solution stability.

For every original view $v$, we generate two separate graph structures: one utilizing the KNN approach, denoted as $\overline{\mathbf{A}}$, and the other employing a semi-supervised method, represented by $\mathbf{S}$. Thus, we artificially generate $2V$ views by constructing two distinct graphs for each of the original $V$ views.

$\mathbf{G}_w$ denotes the stacked matrix views of the $2V$ graphs constructed in both ways, where $w = 1, \ldots, 2V$:

\begin{equation}\mathbf{G}_w=
\begin{cases}
\overline{\mathbf{A}}_{\frac{w+1}{2}}, & \mathrm{if} \; \; w\mathrm{~is~odd}, \\ \\
\mathbf{S}_{\frac{w}{2}}, & \mathrm{if} \; \; w\mathrm{~is~even.}
\end{cases}
\label{eq:fusedG}
\end{equation}

where $w$ denotes the $w$-th view among a total of $2V$ views. Consequently, it is necessary to construct $2V$ feature matrices from the original $V$ feature matrices:

\begin{equation}\overline{\mathbf{X}}_w=
\begin{cases}
\mathbf{X}_{\frac{w+1}{2}}, & \mathrm{if} \;\;w\mathrm{~is~odd}, \\ \\
\mathbf{X}_{\frac{w}{2}}, & \mathrm{if} \;\;w\mathrm{~is~even}.
\end{cases}\end{equation}

The integration of KNN graphs and semi-supervised graphs $S_v$ in graph construction enables the intrinsic structure of data to be captured from multiple perspectives. The KNN graph focuses on modeling local geometric relationships based on sample proximity, reflecting the natural similarity between data points. In contrast, the semi-graph incorporates label information and feature transformations to enhance the graph’s discriminative power, aligning the graph structure more closely with class distributions. These two methods complement each other by preserving unsupervised neighborhood information while integrating semi-supervised signals, effectively reducing the impact of noise and outliers.

\normalsize\textit{\subsubsection{Multi-GCN Classification}}
As the number of views has been increased from $V$ to $2V$, the classification process now employs $2V$ GCNs. Here, the subscript $w$ refers to the $w$-th classifier out of the $2V$ GCNs.

The detailed process of forward propagation for the $w$-th view is as follows:
\begin{equation}
\mathbf{H}_w^{(l+1)}=\sigma\bigl(\mathbf{G}_w\mathbf{H}_w^{(l)}\mathbf{W}_w^{(l)}\bigr)
\label{eq:Hv}
\end{equation}

In the $l$-th layer, the learnable parameter associated with the $w$-th view is represented by $\mathbf{w}_w^{(l)}$, while $\sigma$ denotes a nonlinear activation function. For each $w$-th view, the output of the $l$-th layer is $\mathbf{H}_w^{(l+1)} \in \mathbb{R}^{n \times d_w^{(l+1)}}$, where the initial input is set as ${\mathbf{H}}_w^{(0)} = \overline{\mathbf{X}}_w$. At the output layer, we fuse multi-view neighborhood information via the fused graph matrix $\mathbf{G}_f$. In this way, the final layer of each GCN operates on the fused graph:

\begin{equation}
{\mathbf{G}}_f=\frac{1}{2V}\sum_{w=1}^{2V}{\mathbf{G}}_w
\label{eq:fusedG_f}
\end{equation}

The embedding $\mathbf{Z}_w$ for the $w$-th view in a three-layer model is obtained using the following computation:
\begin{equation}
\mathbf{Z}_w=softmax(\mathbf{G}_{f}\sigma\big(\mathbf{G}_{w}\sigma\big(\mathbf{G}_{w}\overline{\mathbf{X}}_{w}\mathbf{W}_{w}^{(0)}\big)\mathbf{W}_{w}^{(1)}\big)\mathbf{W}_{w}^{(2)})
\label{eq:Z_v}
\end{equation}

\normalsize\textit{\subsubsection{Architecture Output: Soft Voting}}

According to Eq.\ref{eq:Z_v}, $\mathbf{Z}_w$ denotes the output probability generated by the $w$-th GCN. Each GCN serves as an independent classifier. Integrating multiple classifiers generally enhances generalization, often yielding better results than relying on a single classifier. Soft voting over the predictions of $2V$ GCNs classifiers yields the final classification:

\begin{equation}\mathbf{Z}=\frac{1}{2V}\sum_{v=1}^{2V}\mathbf{Z}_{w}
\label{eq:Z}
\end{equation}

Here, $\mathbf{Z}_w$ represents the output from the $w$-th classifier. Specifically, for odd values of $w$, $\mathbf{Z}_w$ corresponds to the classifier associated with the $((w+1)/2)$-th view, constructed using the K-nearest neighbors graph. For even values of $w$, $\mathbf{Z}_w$ denotes the output of the classifier for the $(w/2)$-th view, which is built upon the semi-supervised graph construction method. By integrating classifiers developed using different graph construction strategies, the model is able to more effectively capture the complex and heterogeneous data distributions inherent in multi-view settings. Moreover, since individual classifiers may be susceptible to noise and outliers, employing a soft voting mechanism across multiple classifiers mitigates their impact, thereby enhancing the robustness and reliability of the overall model.

\normalsize\textit{\subsection{Loss Function}}
x

Without loss of generality, we assume that each view employs a 3-layer GCN architecture. For the $w$-th view, we define the feature embedding as the output of the second layer, denoted by $\mathbf{H}_{w}^{2} \in \mathbb{R}^{n \times d_{w}^{(2)}}$, where each row corresponds to the transformed features of a single sample. The embedding for a sample $\mathbf{x}_{wi}$ is represented as $f(\mathbf{x}_{wi}) \in \mathbb{R}^{d_{w}^{(2)}}$, which maps the sample vector $\mathbf{x}_{wi}$ into a $d_{w}^{(2)}$-dimensional Euclidean space. In our work, the latent representations of the nodes for each view are obtained from the output of the second layer, denoted as $\mathbf{H}^2_w$. This is  given by $\mathbf{H}^2_w =  [ f(\mathbf{x}_{w1}); f(\mathbf{x}_{w2}); \ldots;  f(\mathbf{x}_{wn})]$.

\subsubsection{CE loss}

The multi-view cross-entropy loss being minimized is formulated as:  

\begin{equation}  
\mathcal{L}_{CE} = -\sum_{i \in \Omega} \sum_{j=1}^{c} Y_{ij} \log Z_{ij}  
\label{eq:ce}  
\end{equation}  
where \(\mathbf{Y} \in \mathbb{R}^{|\Omega| \times c}\) represents the label matrix associated to the labeled set \(\Omega\), and \(c\) denotes the total number of classes.

\subsubsection{Supervised Contrastive Loss}
To enhance the discriminative power of feature representations across all GCN branches using labeled samples, the Supervised Contrastive Loss is extended to the multi-view setting as follows:
\begin{equation}
\label{eq:supcon}
\begin{aligned}
\mathcal{L}_{SupCon}=\sum_{w=1}^{2V} \sum_{i \in \Omega}  \frac{-1}{|P(i)|}\sum_{p\in P(i)}\\\log\frac{\exp{(f(x_w^i)\boldsymbol{\cdot}f(x_w^p)/\tau)}}{\sum_{a\in A(i)}\exp{(f(x^i_w)\boldsymbol{\cdot}f(x_w^a)/\tau)}}\end{aligned}
\end{equation}

Here, the symbol $\cdot$ denotes the inner (dot) product, and $\tau \in \mathbb{R}^+$ is a scalar temperature parameter. The set $\Omega$ represents the indices of labeled nodes; for each $i$, $A(i) \equiv \Omega \setminus \{i\}$ denotes the set of all labeled node indices excluding $i$. The set $P(i) \equiv \{ p \in A(i) \mid y_p = y_i \}$ consists of the indices of all positive examples (i.e., those with the same label as $i$) distinct from $i$, and $|P(i)|$ denotes its cardinality.

\subsubsection{Self-Supervised Contrastive Loss}

The objective is to enhance the latent representation of each unlabeled sample by contrasting the outputs of the GCN branch constructed using the semi-supervised graph. This loss is given by:

\begin{equation}\begin{aligned}\mathcal{L}_{SelfCon}=- \frac{1}{V (V - 1)}\sum_{i\in U} \sum_{w=1}^{V} \sum_{w^{\prime}=1, \neq w}^{V} \\ \log\frac{\exp\left(f(x_w^i)\cdot f(x_{w^{\prime}}^i)/\tau\right)}{\sum_{a\in A(i)}\exp\left(f(x_w^i)\cdot f(x_{w^{\prime}}^a)/\tau\right)}
\label{eq:supcon-self}\end{aligned}
\end{equation}
where $A(i)\equiv U \setminus \{i\}$ and $U$ is the set of unlabeled samples. 

The embedding $f(x^i_{w})$ is called the $anchor$ of $i$ in view $w$, and $f(x^i_{w'})$ in view $w'$ is another augmented embedding of the same sample $i$ called the positive and the other samples are called the negatives.

\subsubsection{Pseudo-Labels Loss}

Eq.\eqref{eq:ce} and Eq.\eqref{eq:supcon} calculate the CE loss and SupCon loss for the labeled samples. To increase the amount of supervision during training, we can also use the predicted labels of unlabeled samples.
    
By doing so, we can generate a corresponding loss for the unlabeled portion of the data.
Specifically, we use Eq.\eqref{eq:ce} to compute the CE loss for the unlabeled samples, utilizing Pseudo Labels:
\begin{equation}\mathcal{L}_{CE-pseudo}=-\sum_{i\in U_{pred}}\sum_{j=1}^{c}Y_{ij}^{pred} \ln Z_{ij}
\label{eq:ce-pseudo}
\end{equation}
where \textcolor{black}{each element in the prediction matrix \( \mathbf{Y} ^{{pred}} \) represents a continuous probability estimate within the closed interval \([0, 1]\), quantifying the model's confidence that a given sample belongs to a specific class.} $U_{pred}$ is a subset of unlabeled data representing the top \textcolor{black}{$N$} unlabeled samples with the highest confidence. The matrix of pseudo labels $\mathbf{Y} ^{pred}$ is composed of the output predictions from the deep network obtained during the previous epoch.

The Combined Cross-Entropy Loss will be:
\begin{equation}\mathcal{L}_{CE-combine} = \mathcal{L}_{CE} + \lambda_1 \cdot\mathcal{L}_{CE-pseudo}
\label{eq:ce-combine}
\end{equation}

This loss integrates the standard cross-entropy loss computed on labeled samples with the cross-entropy loss calculated on pseudo-labeled samples, enabling the model to be jointly optimized for classification accuracy across both true labeled and pseudo-labeled data.

We can also use pseudo-labels to enhance the ability to discriminate between samples using the following formula:

\begin{equation}\begin{aligned}
\mathcal{L}_{SupCon-pseudo}=\sum_{w=1}^{2V}\sum_{i\in U_{pred}}\frac{-1}{|P^{pred}(i)|}\sum_{p\in P^{pred}(i)}\\ \log\frac{\exp{(f(x_{w}^{i})\cdot f(x_{w}^{p})/\tau)}}{\sum_{a\in A^{pred}(i)}\exp{(f(x_{w}^{i})\cdot f(x_{w}^{a})/\tau)}}\end{aligned}
\label{eq:supcon-pseudo}
\end{equation}

Here $A^{pred}(i)\equiv U_{pred}\setminus\{i\}$, 
$P^{pred}(i)\equiv\{p\in A^{pred}(i):{\boldsymbol{y}}_{p}=
{\boldsymbol{y}}_{i}\}$ is the indexed set of all pseudo positives with the same label as $i$.

The Combined SupCon Loss will be:
\begin{equation}\mathcal{L}_{SupCon-combine} = \mathcal{L}_{SupCon} + \lambda_2 \cdot\mathcal{L}_{SupCon-pseudo}
\label{eq:supcon-combine}
\end{equation}
where the combined loss integrates the supervised contrastive learning loss on labeled data with the supervised contrastive loss on pseudo-labeled data, enabling the model to learn more discriminative feature representations from both true and pseudo labels.

\textcolor{black}{In Eq. (\ref{eq:supcon-self}), the SelfCon loss leverages a semi-supervised graph to form positive sample pairs for the same instance, while all other sample pairs are treated as negative. However, this approach may inadvertently misclassify same-class pairs as false negatives. To mitigate this issue, we introduce pseudo-labeling within the SelfCon loss, increasing the number of positive pairs while reducing false negatives. This refinement enables the model to learn more discriminative features. Consequently, the pseudo-labeling-enhanced SelfCon loss is formulated as:}

\begin{equation}\begin{aligned}
\mathcal{L}_{SelfCon}=-\frac{1}{V (V - 1)}
\sum_{i\in U} \sum_{w=1}^{V} \sum_{w^{\prime}=1, \neq w}^{V}
\frac{1}{|P^{pred}(i)|}\sum_{p\in P^{pred}(i)}\ \\ \log\frac{\exp\left(f(x_w^i)\cdot f(x_{w^{\prime}}^p)/\tau\right)}{\sum_{a\in A(i)}\exp\left(f(x_w^i)\cdot f(x_{w^{\prime}}^a)/\tau\right)}
\label{eq:supcon-self-pseudo}\end{aligned}
\end{equation}

where $P^{pred}(i)\equiv\{p\in U_{pred} :\tilde{\boldsymbol{y}}_{p}=
\tilde{\boldsymbol{y}}_{i}\}$ is the indexed set of all pseudo positives with the same label as $i$(including $i$). 

\subsubsection{Proposed Loss}

The overall loss function utilized in our proposed approach is defined as follows:

\begin{equation}\begin{aligned}
\mathcal{L}_{total}=\mathcal{L}_{CE-combine}+ \mathcal{L}_{SupCon-combine}+ \mathcal{L}_{SelfCon}
\label{eq:total}\end{aligned}
\end{equation}
The proposed loss function is a comprehensive formulation that integrates multiple components, including classification losses and both intra-view and inter-view contrastive losses. These losses are constructed based on both the ground-truth labels and the predicted labels of unlabeled samples, enabling the model to learn discriminative representations while aligning views effectively. This combined objective allows the GCN branches to be trained in an end-to-end fashion, leveraging different levels of supervision.

For the first term,  $\mathcal{L}_{CE-combine}$ defines the cross-entropy loss function for supervised learning on both genuine and pseudo-labeled samples, minimizing classification error by comparing the predictive distribution with the corresponding label distributions.

For the second term, $\mathcal{L}_{SupCon-combine}$ combines supervised contrastive loss on genuine labeled samples and pseudo-labeled samples, aiming to improve feature representation by enforcing stronger intra-class compactness and better discrimination across both true and pseudo labels.

The third term, \(\mathcal{L}_{SelfCon}\), denotes the self-supervised contrastive loss, which is primarily applied to unlabeled data to facilitate the learning of more generalized feature representations. Specifically, it compares augmented samples derived from the same data instance but constructed as different views using the semi-supervised graph, thereby encouraging the model to obtain consistent representations across these views.

\begin{figure}[htp]
     \begin{subfigure}[b]{0.5\textwidth}
         \centering
         \includegraphics[width=0.5\linewidth]{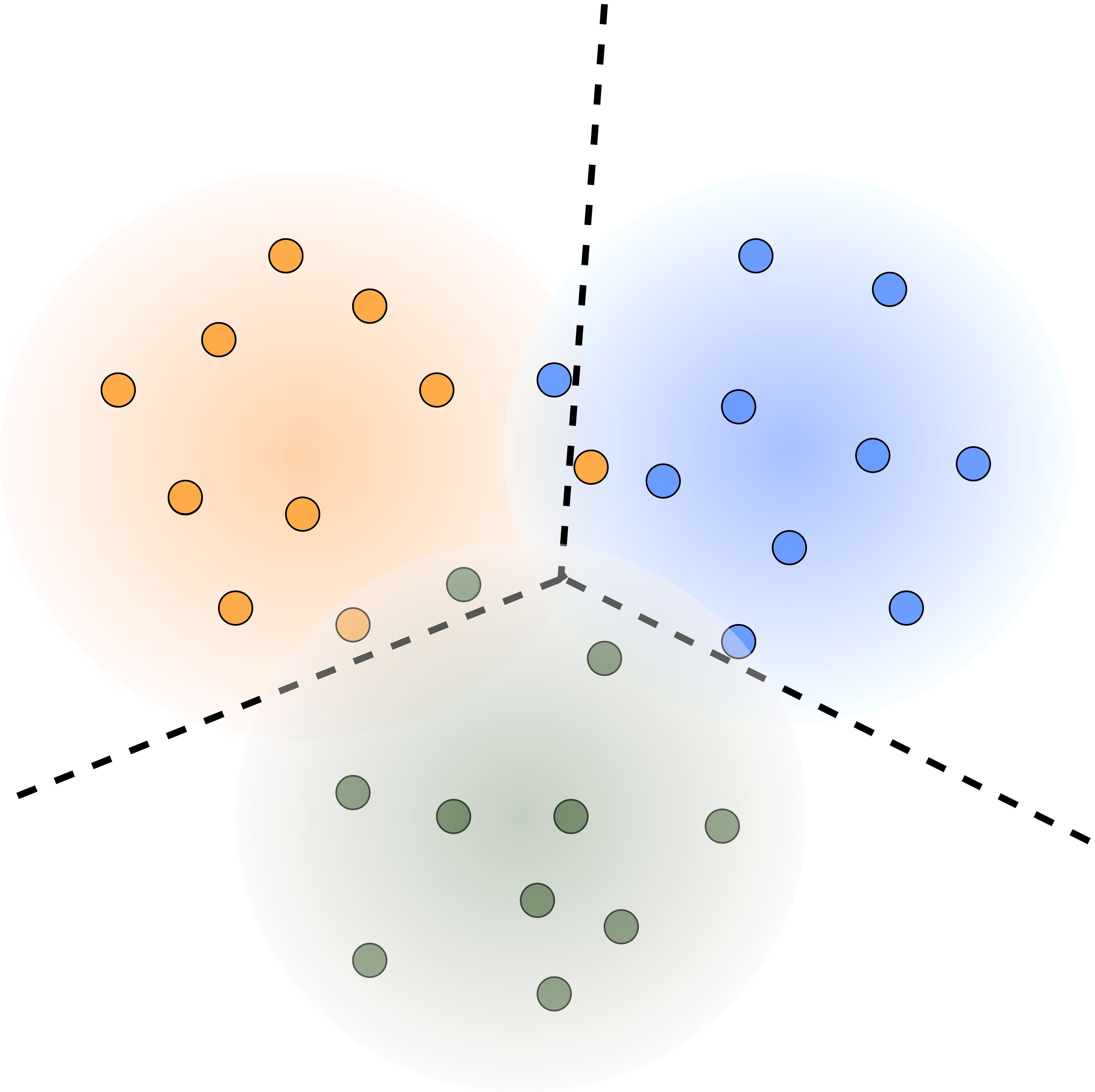}
         \caption{cross-entropy loss}
     \end{subfigure}
     \begin{subfigure}[b]{0.5\textwidth}
         \centering
         \includegraphics[width=0.5\linewidth]{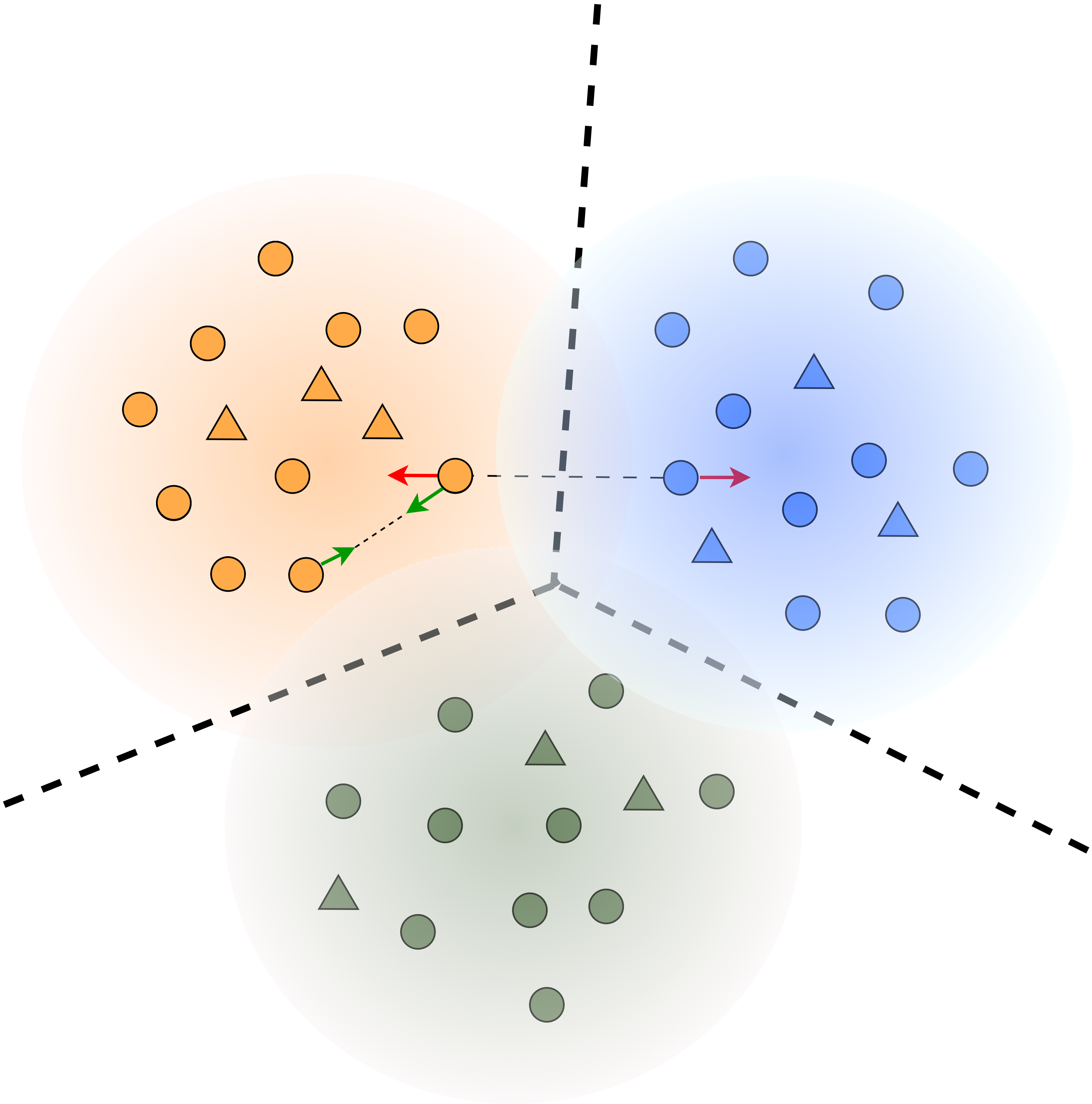}
         \caption{cross-entropy loss + SupCon loss + pseudo label}
     \end{subfigure}  
     \caption{The schematic depicts the feature distributions obtained by optimizing with cross-entropy loss versus the joint use of SupCon and cross-entropy losses. The cross-entropy classifier defines a decision boundary, while SupCon loss pulls features of the same class closer and pushes features of different classes apart, improving class separation. The inclusion of pseudo-labeled samples (triangles) further enhances this effect, making the distance of intra-classes more closer and the decision boundary more accurate through the joint influence of pseudo-labeling and SupCon loss.}
     
     \label{fig:ce_SupCon}
\end{figure}

Algorithm~\ref{algo:MV-SupGCN} presents a summary of the method’s main stages.

\begin{algorithm}[h]
\caption{MV-SupGCN}
\textbf{Input:} Multi-view data $\{\mathbf{X}_1, \ldots, \mathbf{X}_v\}$, semi-supervised information $Y$, Number of iterations $e_{max}$, Number of GCN network layers $L$  (it is set to 3) \\
\textbf{Output:} The fused embedding $\mathbf{Z}$.
\begin{algorithmic}[1]
\STATE Generate adjacency matrices $\{\mathbf{\overline{A}}_1, \ldots, \mathbf{\overline{A}}_v\}$with \textcolor{black}{Eq.\eqref{eq:tildeAv}};
\STATE Generate adjacency matrices $\{\mathbf{{S}}_1, \ldots, \mathbf{{S}}_v\}$with \textcolor{black}{Eq.\eqref{eq:SGraph}};
\STATE Generate the $2V$ adjacency matrices $\mathbf{G}_w$ with \textcolor{black}{Eq.\eqref{eq:fusedG}};
\STATE Calculate the fused graph $\mathbf{G}_f$ with \textcolor{black}{Eq.\eqref{eq:fusedG_f}};
\FOR{$e=1$ to $e_{max}$}
    \FOR{$w = 0$ to $2V$}
        \FOR{$l = 1$ to $L-1$} 
            \STATE Compute $\mathbf{H}_w^{(l)}$ with \textcolor{black}{Eq.\eqref{eq:Hv}};
            \IF{$l = L-1$}
                \STATE Compute $\mathcal{L}_{\text{SupCon-combine}}$ with \textcolor{black}{Eq.\eqref{eq:supcon-combine}};
                
                \STATE Compute $\mathcal{L}_{\text{SelfCon}}$ with \textcolor{black}{Eq.\eqref{eq:supcon-self-pseudo}};
            \ENDIF
        \ENDFOR
    \STATE Calculate the w-th view output embedding $\mathbf{Z}_w$ with \textcolor{black}{Eq.\eqref{eq:Z_v}}
    \ENDFOR
    \STATE Calculate the embedding $\mathbf{Z}$ with \textcolor{black}{Eq.\eqref{eq:Z}};
    \STATE Compute $\mathcal{L}_{\text{CE-combine}}$ with \textcolor{black}{Eq.\eqref{eq:ce-combine}};
    
    \STATE Update parameter with backward propagation using the total loss $\mathcal{L}_{\text{total}}$ according to \textcolor{black}{Eq.\eqref{eq:total}};
\ENDFOR
\STATE \textbf{return} The fused representation $\mathbf{Z}$;
\end{algorithmic}
\label{algo:MV-SupGCN}
\end{algorithm}

{\color{black}

\normalsize\textit{\subsection{Computational Complexity Analysis}}

The method is divided into two primary components: graph construction and GCN training. For the construction of each semi-supervised view $\mathbf{S}_v$, we apply a shallow learning approach that incorporates a linear transformation. The computational cost for constructing a semi-supervised view is ${\cal {O}}(n^3)$, where $n$ represents the number of samples. Without loss of generality, we assume that the feature dimensions across all views are identical and denoted by $d$. The time complexity of the KNN algorithm for graph construction is ${\cal {O}}(n^2 \cdot d)$, where $d$ refers to the dimensionality of the samples. Therefore, the total time complexity for constructing $V$ semi-supervised and KNN graphs is ${\cal {O}}(V\, (n^3+n^2 \cdot d))$.

In a 3-layer GCN network structure, the computational complexity of the first layer of the GCN is ${\cal {O}}(n^2\cdot d+n\cdot d\cdot d_{1})$, the second layer is ${\cal {O}}(n^2\cdot d_{1}+n\cdot d_{1}\cdot d_{2})$ and the last layer is  ${\cal {O}}(n^2\cdot d_{2} +n\cdot d_{2}\cdot c)$, where $n$ is the number of samples, $d_{1}$ and $d_{2}$ are the dimensions of the hidden layers, $c$ is the number of classes. An epoch of the supervised contrastive loss function has a runtime complexity of ${\cal {O}}(n_l^2)$ where $n_l$ is the number of labeled samples. Apart from this, the time complexity of the self-supervised contrastive loss with a single view is ${\cal {O}}((V-1)\cdot n_u^2)$. where $n_u$ is the number of unlabeled samples.

Thus, the total time complexity of training $V$ 3-layer GCNs is ${\cal {O}}(T \, V [n^2(d+d_{1}+d_{2})+n(d\cdot d_{1}+d_{1}\cdot d_{2}+d_{2}\cdot c)+V\cdot n^2])$, $T$ denotes the number of iterations. By combining the graph construction method and the GCN training method, the total time complexity of the proposed method is ${\cal {O}}(V[n^3+n^2 \cdot d+T[n^2(d+d_{1}+d_{2})+n\cdot(d\cdot d_{1}+d_{1}\cdot d_{2}+d_{2}\cdot c)+V\cdot n^2]])$, it can be simplified as ${\cal {O}}(V[n^3+T[n^2\cdot d+n\cdot d^2 + V\cdot n^2]])$.} \\

\section{Experimental Setup}
\label{sec:experimentalsetup}

This section comprehensively details the experimental setup, encompassing the datasets, comparative methods, parameter settings, and training configurations.

\normalsize\textit{\subsection{Datasets}}

We assess the effectiveness of MV-SupGCN on seven widely recognized benchmark datasets: 3Sources, Citeseer, GRAZ02, MNIST, NGs, Out-Scene, and NoisyMNIST. All selected datasets are designed for multi-view learning and include various types of samples, ensuring a thorough evaluation of the model’s generalizability. Table \ref{tab:datasets_statistics} presents a summary of the key properties of these datasets, with further details provided below:

\begin{table}[h] 
\centering
\caption{Dataset details.}  
\label{tab:datasets_statistics}
\renewcommand{\arraystretch}{1.5}
\begin{tabular}{|l|r|c|r|c|l|}  
\toprule  
Dataset Name& \ \#Samples& \ \#Views& \ Feature Dimensions & \ \#Classes &\ Data type\\
\midrule  
3Sources & 169 & 3 & 3,560/3,631/3,068 & 6 & Textual dataset\\
Citeseer & 3,312 & 2 & 3,703/3,312 & 6 & Textual dataset \\
GRAZ02 & 1,476 & 6 & 512/32/256/500/500/680 & 4 & Object images \\ 
MNIST & 2,000 & 3 & 30/9/9 & 10 & Digit dataset \\ 
NGs & 500 & 3 & 2,000/2,000/2,000 & 5 & Textual dataset\\
Out-Scene & 2,688 & 4 &512/432/256/48 & 8 & Scene image\\
\textcolor{black}{NoisyMNIST} & \textcolor{black}{10000} & \textcolor{black}{3} & \textcolor{black}{784/784/784} & \textcolor{black}{10} & \textcolor{black}{Digit dataset} \\
\bottomrule
\end{tabular}
\end{table}

\textbf{3Sources} is a text dataset comprising news articles on common topics collected from three different sources, each offering distinct feature representations. Among the collected articles, 169 were covered by all three sources. Each article was manually assigned one or more of six topical categories: business, entertainment, health, politics, sport, and technology.

\textbf{Citeseer} dataset contains 3,312 scientific publications, each assigned to one of six categories. Following the approach outlined in \cite{wu2023interpretable}, we represent each document using two views: a 3,703-dimensional bag-of-words vector to capture textual content, and a 3,312-dimensional vector to encode citation relationships among documents.

\textbf{GRAZ02} dataset is used for object categorization and contains images from four categories: bicycles, people, cars, and a category without these objects. Six types of features are extracted from the images: SIFT, SURF, GIST, LBP, PHOG, and WT.

\textbf{MNIST} serves as a widely-used benchmark dataset for handwritten digit recognition. For this work, three types of features are derived from the original data: 30-D IsoProjection, 9-D LDA, and 9-D NPE features.

\textbf{NGs} comprises 500 newsgroup documents, which are divided into five classes. Each document is pre-processed using three different methods to generate three distinct views per sample, with each view characterized by a 2000-dimensional feature representation.

\textbf{Out-Scene} is an image dataset comprising 2,688 samples distributed across eight distinct categories. It includes 512-D GIST, 432-D Color moment, 256-D HOG, and 48-D LBP.

\textcolor{black}{\textbf{NoisyMNIST} is derived from the original MNIST handwritten digit dataset by applying three types of corruption: additive white Gaussian noise, motion blur, and a combination of additive white Gaussian noise with reduced contrast.  Each image is 28×28 pixels, flattened into a 784-dimensional vector, and the labels are one-hot encoded vectors indicating digits from 0 to 9.}

\normalsize\textit{\subsection{Compared Methods}}
\
We evaluated our method by comparing it with several SOTA multi-view semi-supervised algorithms. This comparative study includes methods such as MVCGL \cite{ZIRAKI2022174}, DSRL \cite{9439159},  GCN-fusion \cite{kipf2016semi}, SSGC-fusion \cite{2021Simple}, Co-GCN \cite{2020Co}, ECMGD \cite{10.1145/3664647.3681258}, DG-MVL \cite{LU2025102661}, IMvGCN \cite{wu2023interpretable}, GEGCN\cite{10462517} and MV-TriGCN\cite{XIAO2025103420}.  MVCGL is classified as a shallow model, while the other models are classified as deep models. It is noteworthy that GCN-fusion, Co-GCN, ECMGD, IMvGCN, GEGCN, and MV-TriGCN employ GCN as their backbone network. 

Among them, since GCN and SSGC are unable to process multi-view data directly, we have developed the GCN-fusion and SSGC-fusion variants to address this limitation. \textcolor{black}{Additionally, for all the graph-based methods used as comparison baselines, we use the same K-Nearest Neighbors approach for graph construction. This ensures that the adjacency matrices are generated consistently across all methods, providing a uniform and fair basis for comparison.} The following content introduces these algorithms and their implementation details:

\textbf{MVCGL} introduces the Multiple-view consistent graph construction approach. This approach constructs a unified graph and simultaneously propagates labels to unlabeled samples. We use the default settings for MVCGL and set the hyperparameters to: $p$ = 2, $\eta$ = 5, $\gamma$ = 100, $\mu$ = 0.003, and $\alpha$ = 0.003.

\textbf{DSRL} introduces a deep learning model that adaptively learns sparse regularizers from data, making it more effective at extracting meaningful features. In this method, the architecture uses a fixed number of 10 layers, and the learning rate is consistently set to 0.2.

\textbf{GCN-fusion} performs GCN $V$ times. In each run, the input features, $\mathbf{X}_v$, are used to generate the corresponding adjacency matrix $\mathbf{A}_v$. This process yields a total of $V$ outputs, designated as $\mathbf{Z}_v$. The learning rate is fixed at 0.01 and the output of GCN-fusion is calculated as the mean of outputs from each view: $\mathbf{Z} = \frac{1}{V} \sum_{v=1}^{V} \mathbf{Z}_v$.

\textbf{SSGC-fusion} utilizes linear classifiers and demonstrates strong performance on both text and node classification benchmarks. In all experiments, the learning rate was fixed at 0.001. For each view $v$, the SSGC model generates an output representation $\mathbf{Z}_v$ for classification. Consistent with the strategy adopted in the GCN-fusion framework, the final representation produced by SSGC-fusion is obtained by taking the average over the outputs from all individual views, formulated as $\mathbf{Z} = \frac{1}{V} \sum_{v=1}^{V} \mathbf{Z}_v$.

\textbf{Co-GCN} introduces a novel multi-view semi-supervised learning approach that adaptively leverages graph information from multiple views through the use of combined Laplacians. The learning rate is set to 0.001, and the number of neighbors for graph construction is set to 10.

\textbf{IMvGCN} is an interpretable multi-view graph convolutional network framework that effectively integrates feature and topological information from multiple views by combining reconstruction error and Laplacian embedding, introducing an orthogonal normalization method. For all datasets, the hyper-parameter $\lambda$ is fixed at 0.5 and the balance hyper-parameter $\alpha$ is selected from the range $\{1\times10^{-4}, 1\times10^{-5},1\times10^{-6}\}$.

\textbf{GEGCN} proposes a generative essential graph convolutional network that effectively enhances multi-view semi-supervised classification by integrating multi-graph consistency and complementarity extraction, graph refinement, and downstream task optimization within a unified framework. The maximum number of epochs is set to 700 and the learning rate is set to 0.01.

\textcolor{black}{\textbf{ECMGD}{ introduces an energy-constrained multi-view graph diffusion framework that enables unified feature propagation across and within views, ensuring globally consistent graph embeddings and significantly improving multi-view learning performance.
}}

\textcolor{black}{\textbf{DG-MVL}{ is a divergence-guided multi-view learning framework that leverages multi-view autoencoders to simultaneously disentangle and optimize common and unique features from different views, producing comprehensive representations that enhance downstream task performance.
}}

\textbf{MV-TriGCN} introduces an enhanced triplet loss, flexible multi-view graph construction, and a stepwise training strategy to effectively improve representation learning and performance in multi-view semi-supervised graph convolutional networks. All parameters of the algorithm are configured in accordance with the original paper to ensure consistency in the experimental setup.

\normalsize\textit{\subsection{Training Setting}}

Each of the ten models was evaluated by performing ten independent runs on every dataset. For each run, the dataset was randomly partitioned into ten distinct training and testing splits. All models were trained and tested on these predetermined splits. The mean accuracy and standard deviation were calculated based on the results obtained from these ten divisions. To ensure the fairness and consistency of the experimental comparisons, all models were assessed using the same set of ten data splits.

\textcolor{black}{In each run, only the same labeled samples assigned to the training split were used for both semi-supervised graph construction and model training. The graph was built exclusively with these labeled samples prior to model training and remained fixed throughout the training and testing phases. No label information from the remaining samples was utilized at any stage. This design ensures that the evaluation is conducted under consistent label constraints across all models, with no information from the test split incorporated into graph construction or model learning, thereby guaranteeing the fairness and rigor of the experimental comparisons.}

\begin{table}[h] 
\centering
\caption{Optimal parameter settings for different datasets.}  
\label{tab:datasets_optimal}
\renewcommand{\arraystretch}{1.5}
\begin{tabular}{|l|c|c|c|c|}  
\toprule  
Dataset& \ \makecell{First hidden \\layer reduction}& \ \makecell{Second\\hidden layer\\ features}& \ $e_{max}$  & \ \makecell{TOP $N$ \\pseudo-label \\ ratio}\\
\midrule  
3Sources & 4 & 70  & 50 & 10\% \\
Citeseer & 8 & 400  & 20 & 50\% \\
GRAZ02 & 4 & 30 & 100 & 20\% \\ 
MNIST & 2 & 20  & 100 & 20\% \\ 
NGs & 16 & 30  & 100 & 20\% \\
Out-Scene & 2 & 100 & 100 & 20\% \\
\bottomrule
\end{tabular}
\end{table}

\normalsize\textit{\subsection{Parameter Settings}}

We use the Adam optimizer with a learning rate of $1\times10^{-2}$ to optimize the proposed MV-SupGCN. Each branch in the proposed architecture consists of a 3-layer GCN. During the graph construction step, as described in Eq. (\ref{eq:SGraph}), we adopt the default parameter settings of the MVCGL method. The configuration of the GCN hidden units is based on the original feature dimensionality of each dataset. Specifically, the number of units in the first hidden layer is determined by dividing the initial feature dimension by a scaling factor selected from $\{2, 4, 8, 16\}$, while the number of units in the second hidden layer is selected from a range of 20 to 400. Training stops when the preset maximum iterations $e_{max}$, chosen from $\{20, 50, 100\}$, is reached. During training, in each epoch, the top $N$ unlabeled samples with the highest confidence scores are selected as pseudo-labels for the subsequent epoch. The proportion of top $N$ samples is selected from \{10\%,20\%,50\%\}.The hyper-parameter \(\beta\) was uniformly set to 0.5 across all datasets. Moreover, the \(\tanh(\cdot)\) function was utilized as the activation function. The KNN algorithm, which is employed to construct the adjacency matrix, uses a hyper-parameter \(K\) that ranges from 5 to 40. A summary of the optimal parameter configurations for the MV-SupGCN model is provided in Table~\ref{tab:datasets_optimal}. \\

\begin{table}[h!]
{\color{black}
\centering
\caption{Performance comparison across seven datasets (mean\% ± std.\%) — Accuracy (ACC) / F1 Score (F1) / Precision (P) / Recall (R).}
\renewcommand{\arraystretch}{1}
\setlength{\tabcolsep}{3pt} 

\resizebox{\textwidth}{!}{
\begin{tabular}{llccccccc}
\toprule
Method & Metric & Citeseer & NGs & 3Sources & MNIST & Out-Scene & GRAZ02 & NoisyMNIST \\
\midrule

\multirow{4}{*}{MVCGL \cite{ZIRAKI2022174}} 
& ACC & $57.18\pm3.68$ & $86.67\pm4.39$ & $67.72\pm3.54$ & $83.74\pm1.40$ & $76.04\pm1.27$ & $51.25\pm3.26$ & $90.59 \pm 0.47$ \\
& F1  & $55.41\pm3.36$ & $87.92\pm3.90$ & $50.09\pm8.23$ & $80.67\pm1.80$ & $78.27\pm1.08$ & $54.27 \pm 3.72$ & $91.21 \pm 0.43$ \\
& P   & $63.19\pm1.58$ & $89.83\pm2.34$ & $\mathbf{89.33\pm2.30}$ & $88.11\pm1.51$ & $80.65\pm0.62$ & $\mathbf{59.60\pm2.32}$ & $91.87 \pm 0.43$ \\
& R   & $54.84\pm3.49$ & $87.76\pm4.09$ & $48.51\pm6.90$ & $79.24\pm1.79$ & $77.62\pm1.13$ & $54.12\pm3.18$ & $91.05 \pm 0.44$ \\
\midrule

\multirow{4}{*}{DSRL \cite{9439159}} 
& ACC & $61.68\pm1.54$ & $86.44\pm1.99$ & $70.25\pm2.73$ & $74.63\pm0.79$ & $67.33\pm1.77$ & $50.02\pm1.25$ & $83.33\pm0.41$ \\
& F1  &$55.47\pm1.78$ &$86.31\pm2.04$& $48.06\pm6.40$ & $69.87\pm1.46$ & $67.32\pm2.12$ & $49.25\pm1.42$ & $75.77\pm0.36$ \\
& P   &$58.90\pm1.88$ & $86.53\pm2.05$ & $65.02\pm7.66$ & $83.17\pm1.79$ & $70.36\pm1.73$ & $53.22\pm1.57$ & $76.17\pm0.35$ \\
& R   &$55.82\pm1.62$ & $86.44\pm1.99$ & $47.20\pm5.29$ & $67.32\pm1.11$ & $66.72\pm1.96$ & $49.13\pm1.35$ & $75.48\pm0.38$ \\
\midrule

\multirow{4}{*}{GCN-fusion \cite{kipf2016semi}} 
& ACC & $68.13\pm0.85$ & $95.87\pm1.01$ & $80.12\pm3.82$ & $90.35\pm0.87$ & $75.45\pm1.51$ & $54.99\pm1.29$ & $89.71\pm0.62$ \\
& F1  & $59.83\pm0.66$ & $95.89\pm0.99$ & $64.32\pm5.43$ & $86.88\pm1.89$ & $75.95\pm1.46$ & $54.75\pm1.25$ & $89.57\pm0.64$ \\
& P   & $63.83\pm4.54$ & $96.02\pm0.97$ & $64.93\pm5.73$ & $90.16\pm1.14$ & $77.10\pm1.19$ & $56.31\pm1.28$ & $89.73\pm0.62$ \\
& R   & $61.45\pm0.57$ & $95.87\pm1.01$ & $65.69\pm5.40$ & $86.19\pm1.53$ & $75.66\pm1.62$ & $54.57\pm1.33$ & $89.56\pm0.64$ \\
\midrule

\multirow{4}{*}{SSGC-fusion \cite{2021Simple}} 
& ACC & $68.01\pm1.14$ & $89.22\pm1.89$ & $79.32\pm2.50$ & $\underline{90.85\pm0.71}$ & $77.13\pm0.98$ & $54.93\pm1.27$ & $89.44\pm0.72$ \\
& F1  & $58.91\pm0.98$ & $89.21\pm2.11$ & $61.75\pm4.22$ & \underline{$88.41\pm1.11$} & $77.52\pm0.92$ & $54.73\pm1.45$ & $89.26\pm0.73$ \\
& P   & $60.10\pm4.92$ & $89.57\pm2.02$ & $67.94\pm2.11$ & $90.62\pm0.78$ & $78.92\pm0.90$ & $55.61\pm1.58$ & $89.42\pm0.73$ \\
& R   & $60.82\pm1.11$ & $89.16\pm2.17$ & $61.72\pm4.63$ & \underline{$87.52\pm1.13$} & $77.13\pm0.96$ & $54.70\pm1.41$ & $89.27\pm0.73$ \\
\midrule

\multirow{4}{*}{Co-GCN \cite{2020Co}} 
& ACC & $64.46\pm1.05$ & $85.58\pm2.81$ & $68.82\pm5.53$ & $86.75\pm1.21$ & $65.69\pm0.93$ & $56.45\pm2.11$ & $88.14\pm0.82$ \\
& F1  & $56.72\pm1.34$ & $85.56\pm2.82$ & $45.38\pm7.42$ & $83.50\pm1.60$ & $65.80\pm1.03$ & $55.65\pm4.71$ & $87.92\pm0.84$ \\
& P   & $64.16\pm4.10$ & $86.02\pm2.40$ & $52.95\pm7.98$ & $86.57\pm1.34$ & $67.22\pm1.55$ & $56.43\pm5.04$ & $88.18\pm0.83$ \\
& R   & $55.56\pm1.17$ & $87.16\pm2.82$ & $56.50\pm7.54$ & $82.49\pm1.53$ & $64.77\pm0.91$ & $42.64\pm4.59$ & $88.76\pm0.82$ \\
\midrule

\multirow{4}{*}{IMvGCN \cite{wu2023interpretable}} 
& ACC & $68.42\pm0.74$ & $96.61\pm0.37$ & $80.50\pm4.35$ & $88.79\pm1.02$ & $72.79\pm2.04$ & $55.52\pm2.03$ & $89.36\pm0.70$ \\
& F1  & $64.40\pm0.54$ & $96.61\pm0.37$ & $71.73\pm5.52$ & $86.32\pm2.25$ & $73.00\pm2.13$ & $54.29\pm2.49$ & $89.17\pm0.72$ \\
& P   & $64.72\pm0.44$ & $96.63\pm0.37$ & $72.65\pm6.77$ & $88.37\pm2.83$ & $73.17\pm2.01$ & $54.68\pm2.36$ & $89.36\pm0.76$ \\
& R   & $64.68\pm0.54$ & $96.61\pm0.37$ & \underline{$74.91\pm5.46$}
 & $85.68\pm1.78$ & $73.35\pm1.99$ & $55.81\pm2.07$ & $89.18\pm0.71$ \\
\midrule

\multirow{4}{*}{GEGCN \cite{10462517}} 
& ACC & $63.44\pm0.91$ & $70.02\pm10.52$ & $65.75\pm5.30$ & $88.81\pm0.86$ & $72.40\pm1.83$ & $52.87\pm1.76$ & $91.60\pm0.24$ \\
& F1  & $56.11\pm1.39$ & $64.53\pm13.60$ & $43.27\pm6.26$ & $86.18\pm2.54$ & $72.70\pm1.72$ & $52.97\pm1.76$ & $91.50\pm0.24$ \\
& P   & $56.74\pm2.21$ & $72.80\pm11.23$ & $49.68\pm10.61$ & $86.99\pm3.00$ & $73.26\pm1.66$ & $53.47\pm1.97$ & $91.60\pm0.21$ \\
& R   & $57.55\pm0.95$ & $70.02\pm10.52$ & $47.62\pm5.74$ & $85.94\pm2.12$ & $72.70\pm1.75$ & $52.93\pm1.65$ & \underline{$91.49\pm0.25$} \\
\midrule

\multirow{4}{*}{ECMGD \cite{10.1145/3664647.3681258}} 

& ACC & $61.73\pm1.49$ & $86.04\pm2.17$ & $57.27\pm4.15$ & $86.40\pm0.63$ & $76.07\pm0.92$ & $56.80\pm2.10$ & $85.05\pm0.39$ \\
& F1  & $54.75\pm1.46$ & $86.20\pm2.18$ & $27.45\pm4.44$ & $83.49\pm1.23$ & $76.38\pm0.93$ & $56.42\pm2.13$ & $84.76\pm0.39$ \\
& P   & $60.22\pm2.54$ & $87.60\pm1.61$ & $40.13\pm10.51$ & $85.53\pm1.16$ & $76.99\pm1.01$ & $56.70\pm2.17$ & $84.95\pm0.40$ \\
& R   & $55.62\pm1.37$ & $86.04\pm2.17$ & $31.20\pm3.42$ & $82.83\pm1.12$ & $76.42\pm0.81$ & $56.77\pm2.15$ & $84.78\pm0.38$ \\
\midrule

\multirow{4}{*}{DG-MVL \cite{LU2025102661}} 
& ACC & $59.18\pm0.78$ & $51.68\pm1.97$ & $54.10\pm2.53$ & $81.60\pm0.42$ & $78.54\pm0.30$ & $57.63\pm0.61$ & $88.35\pm0.22$ \\
& F1  & $53.95\pm0.74$ & $47.96\pm2.55$ & $30.57\pm2.16$ & $77.85\pm0.54$ & $78.97\pm0.29$ & $57.23\pm0.69$ & $88.18\pm0.22$ \\
& P   & $57.49\pm0.45$ & $66.89\pm3.62$ & $40.35\pm3.89$ & $80.45\pm0.53$ & $80.23\pm0.30$ & $58.35\pm0.94$ & $88.41\pm0.21$ \\
& R   & $54.37\pm0.81$ & $51.68\pm1.97$ & $34.58\pm2.10$ & $77.16\pm0.52$ & $78.74\pm0.30$ & $57.72\pm0.60$ & $88.16\pm0.22$ \\
\midrule

\multirow{4}{*}{MV-TriGCN \cite{XIAO2025103420}} 
& ACC & $\underline{69.83\pm0.95}$ & $\underline{97.83\pm0.34}$ & $\underline{83.91\pm3.51}$ & $90.49\pm0.46$ & $\underline{79.99\pm0.81}$ & $\underline{59.04\pm1.60}$ & $\underline{91.63\pm0.77}$ \\
& F1  & \underline{$65.24\pm0.96$} & \underline{$97.83\pm0.34$} & $\mathbf{73.22\pm6.58}$ & $88.35\pm1.06$ & \underline{$80.18\pm0.86$} & \underline{$58.12\pm2.20$} & \underline{$91.51\pm0.81$} \\
& P   & \underline{$65.65\pm0.89$} & \underline{$97.85\pm0.34$} & \underline{$72.76\pm7.54$} & \underline{$91.03\pm0.47$} & \underline{$81.08\pm0.87$} & $58.81\pm2.29$ & \underline{$91.96\pm0.63$} \\
& R   & \underline{$65.32\pm0.96$} & \underline{$97.83\pm0.34$} & $\mathbf{75.25\pm5.74}$ & $87.23\pm0.98$ & \underline{$80.20\pm0.77$} & \underline{$59.48\pm1.56$} & $91.48\pm0.79$ \\
\midrule

\multirow{4}{*}{MV-SupGCN (Ours)} 
& ACC & $\mathbf{71.50\pm0.82}$ & $\mathbf{98.32\pm0.25}$ & $\mathbf{85.65\pm1.19}$ & $\mathbf{91.38\pm0.56}$ & $\mathbf{80.64\pm0.86}$ & $\mathbf{60.07\pm1.64}$ & $\mathbf{92.09\pm0.50}$ \\
& F1  & $\mathbf{66.18\pm0.69}$ & $\mathbf{98.31\pm0.25}$ & $\underline{71.85\pm3.84}$ & $\mathbf{89.79\pm0.84}$ & $\mathbf{80.89\pm0.90}$ & $\mathbf{58.99\pm2.08}$ & $\mathbf{91.98\pm0.54}$ \\
& P   & $\mathbf{67.28\pm1.09}$ & $\mathbf{98.32\pm0.25}$ & $70.55\pm4.78$ & $\mathbf{91.65\pm0.46}$ & $\mathbf{81.68\pm0.75}$ & \underline{$59.54\pm2.14$} & $\mathbf{92.21\pm0.48}$ \\
& R   & $\mathbf{66.42\pm0.60}$ & $\mathbf{98.32\pm0.25}$ & $74.77\pm3.18$ & $\mathbf{88.80\pm0.95}$ & $\mathbf{80.90\pm0.87}$ & $\mathbf{60.23\pm1.53}$ & $\mathbf{92.09\pm0.50}$ \\
\bottomrule
\end{tabular}
}
\label{tab:result}
}
\end{table}

\section{Experimental Results}
\label{sec:experimentalresults}
\normalsize\textit{\subsection{Performance Comparison}}

Table~\ref{tab:result} presents the performance comparison of ten different methods on seven benchmark datasets, where only 5\% of the samples are used as labeled data for supervised learning. For each dataset, the method achieving the top classification accuracy is indicated in bold font, and the second-highest accuracy is underlined. Furthermore, both the average accuracy and the corresponding standard deviation are reported for all datasets.

By evaluating classification accuracy across seven different datasets, the experimental results demonstrate that our proposed method, MV-SupGCN, significantly outperforms the competing methods on all datasets. This showcases its exceptional performance and stability in handling multi-view data and diverse classification tasks. These findings fully highlight the promising prospects and potential value of MV-SupGCN as a multi-view semi-supervised graph convolutional network in practical applications. It is noteworthy that our model demonstrates lower standard deviation on most datasets, indicating greater stability in its performance.

\normalsize\textit{\subsection{Sensitivity to Parameters}}

\begin{figure}[h]
\centering  
     \begin{subfigure}[b]{0.45\textwidth}
         \includegraphics[width=\linewidth]{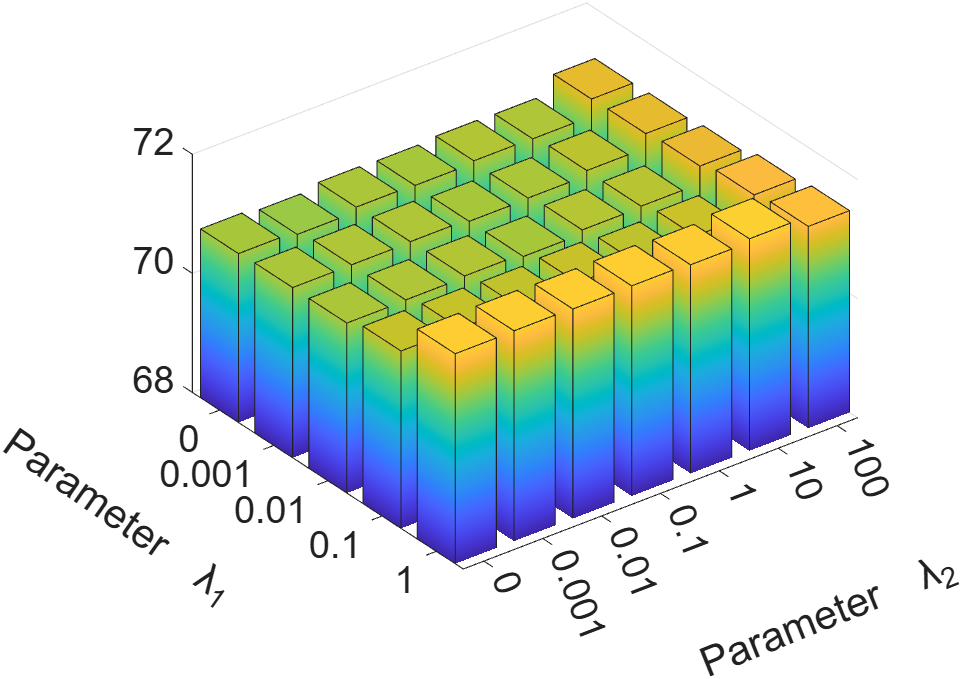}
         \caption{Citeseer}
     \end{subfigure}
     \\
     \begin{subfigure}[b]{0.45\textwidth}
         \includegraphics[width=\linewidth]{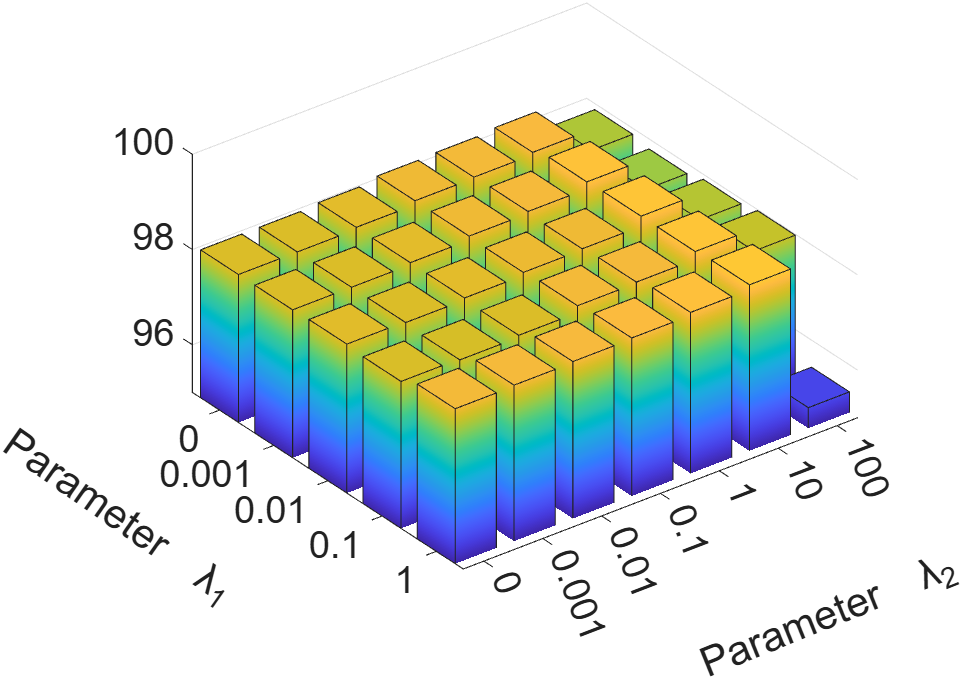}
         \caption{NGs}
     \end{subfigure}

    \caption{Sensitivity analysis of parameters on the Citeseer and NGs datasets.}
    \label{fig:Parameter sensitivity}
\end{figure}

Fig.~\ref{fig:Parameter sensitivity} illustrates the influence of the loss function hyperparameters    $\lambda_1$ and $\lambda_2$     on classification performance using the Citeseer and NGs datasets. We conducted a parameter sensitivity analysis using grid search, which revealed that the hyperparameters impact classification performance. Specifically, as \(\lambda_1\) and \(\lambda_2\) are reduced toward zero, a slight decrease in accuracy is observed. These results suggest that appropriately weighting the pseudo-labels is crucial for enabling more effective learning of the underlying graph structure.

\begin{figure}[h!]
\centering  
     \begin{subfigure}[b]{0.45\textwidth}
         \includegraphics[width=\linewidth]{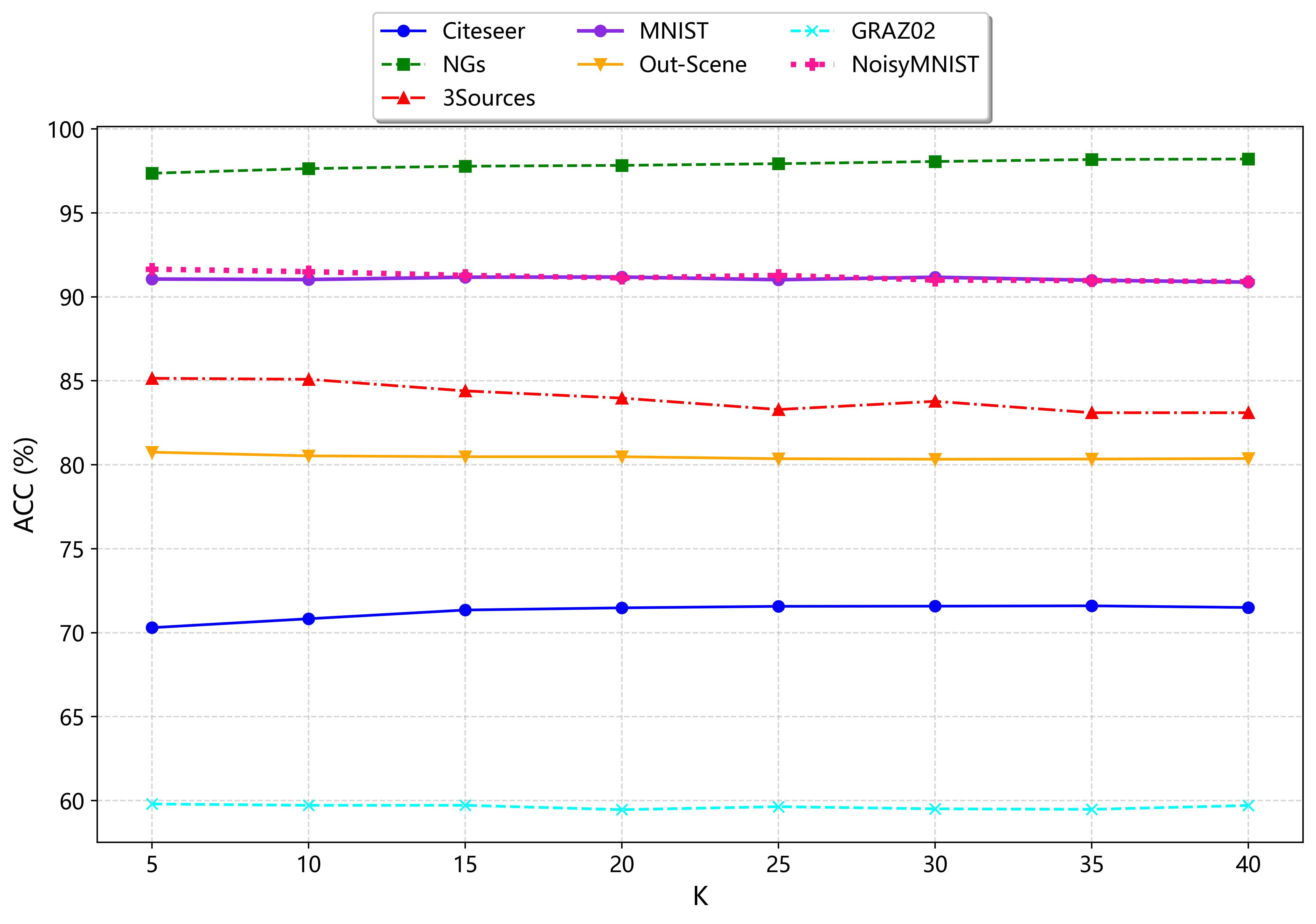}
         \caption{KNN graph construction $K$}
     \end{subfigure}
     \\
     \begin{subfigure}[b]{0.45\textwidth}
         \includegraphics[width=\linewidth]{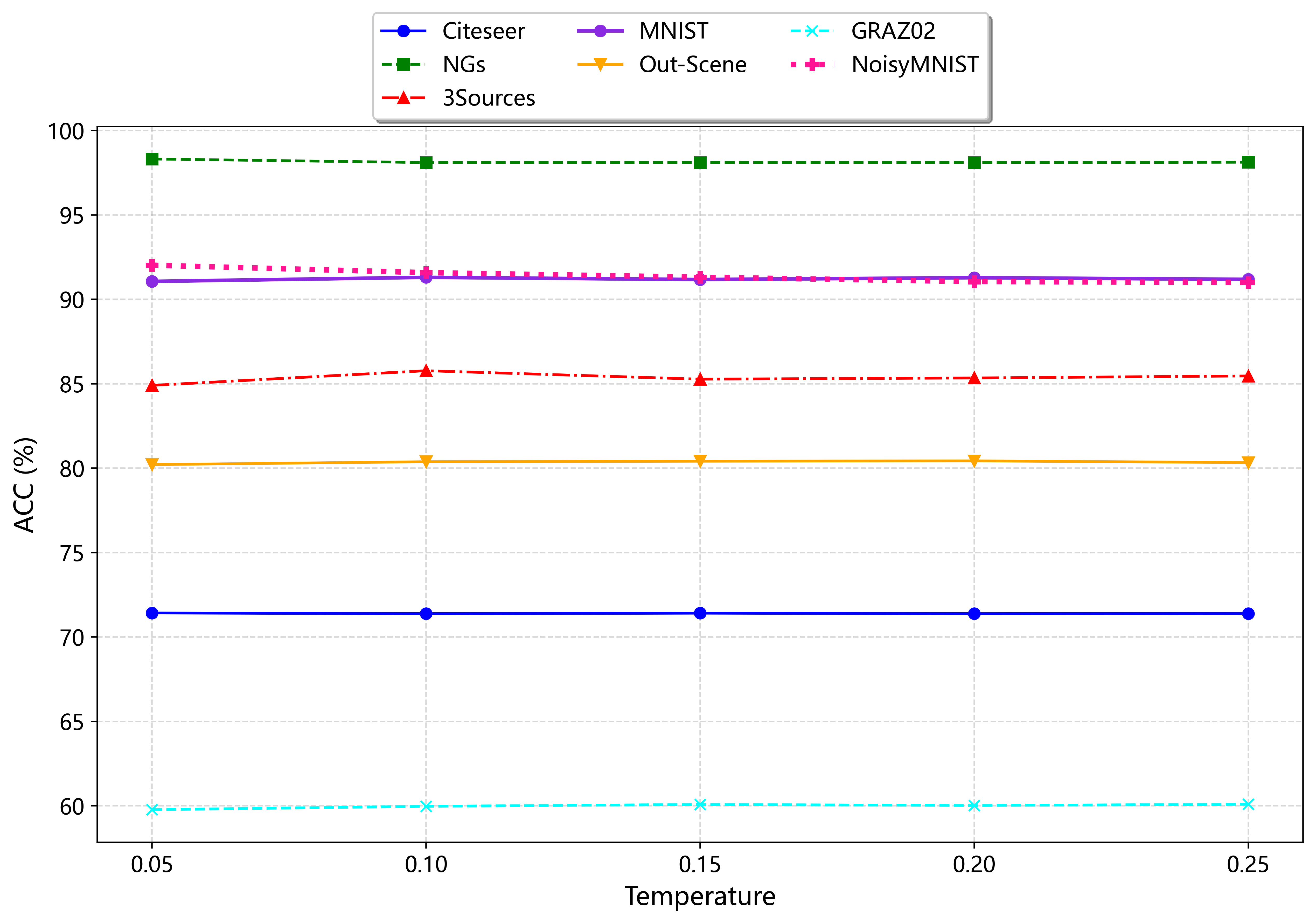}
         \caption{Contrastive loss temperature $\tau$}
     \end{subfigure}
     \\
     \begin{subfigure}[b]{0.45\textwidth}
         \includegraphics[width=\linewidth]{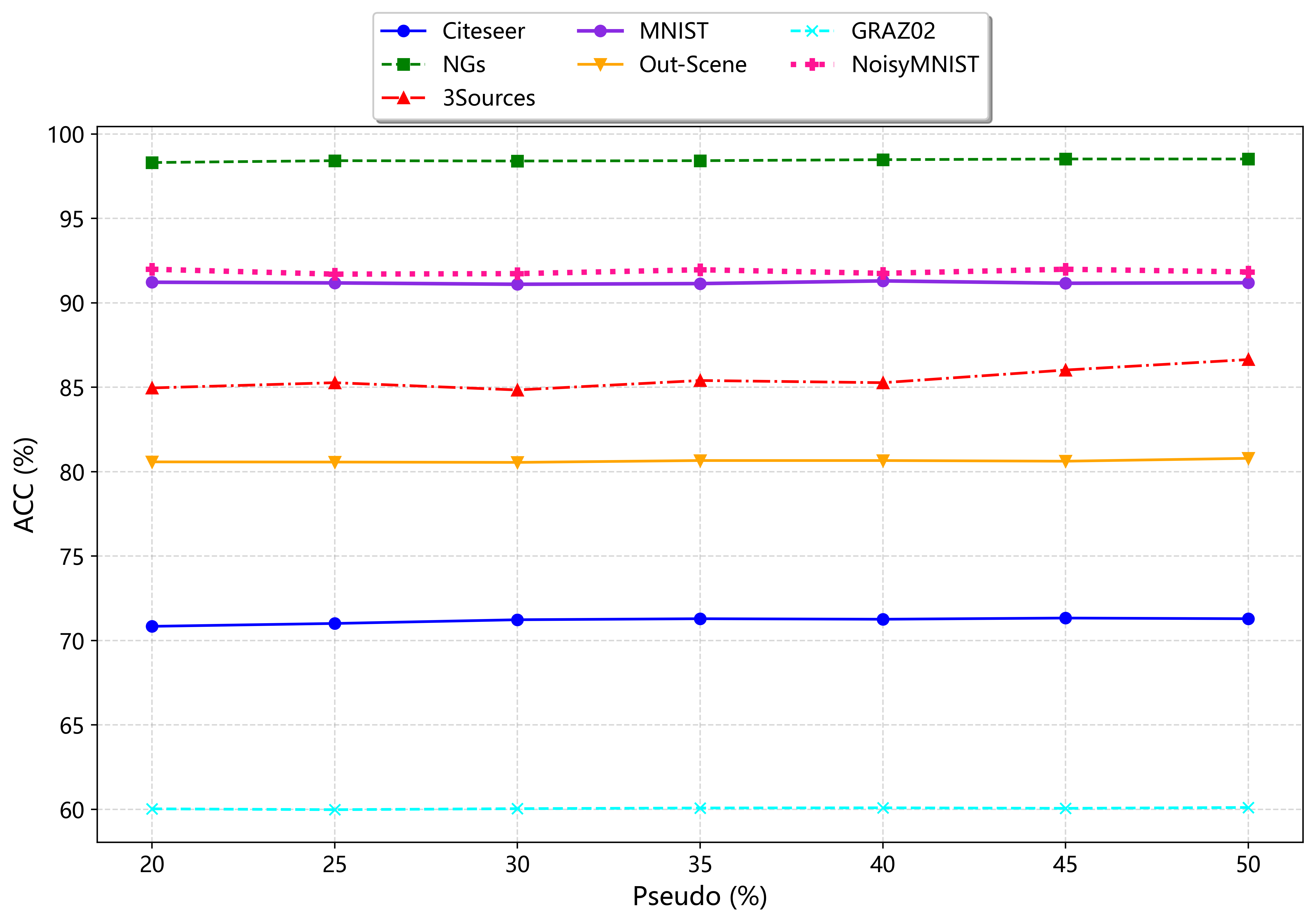}
         \caption{pseudo label ratio}
     \end{subfigure}
    \caption{\textcolor{black}{Impact of key hyperparameters on model robustness.}}
    \label{fig: key parameter sensitivity}
\end{figure}

In greater detail, the parameters $\lambda_1$ and $\lambda_2$ have an interdependent relationship. Fig. \ref{fig:Parameter sensitivity} reveals that when these two parameters are within an appropriate range, the model's performance reaches its optimal point. Therefore, careful selection and tuning of the hyperparameters $\lambda_1$ and $\lambda_2$  can be essential to achieve optimal classification performance.

{\color{black}
To comprehensively evaluate the sensitivity of the model to key hyperparameters, we conducted additional experiments investigating the effects of varying $K$-values(where $K$ denotes the number of neighbors used in the KNN graph construction), temperature settings in the contrastive loss, and pseudo ratio proportions. As shown in Fig.~\ref{fig: key parameter sensitivity}, the model accuracy remains largely stable across a reasonable range of $K$-values and temperature parameters, indicating strong generalization and robustness to these choices. In addition, experiments with different pseudo ratio values demonstrate that the model maintains consistent performance despite variations in the proportion of pseudo-labeled data, further confirming its robustness and adaptability to different pseudo ratio configurations.
}

\normalsize\textit{\subsection{Embedding Visualization}}

We visually compare the original input features $\mathbf{X}$ with the transformed representations $\mathbf{Z}$ produced by the MV-SupGCN method, as defined in Eq. (\ref{eq:Z}). To support this analysis, we use t-SNE to project both the input features from the first view of each dataset and the corresponding outputs. This approach allows us to intuitively observe and contrast the distributions and structural differences between $\mathbf{X}$ and $\mathbf{Z}$.

\begin{figure}[h!]
     \begin{subfigure}[b]{0.23\textwidth}
         \centering
         \includegraphics[width=\linewidth]{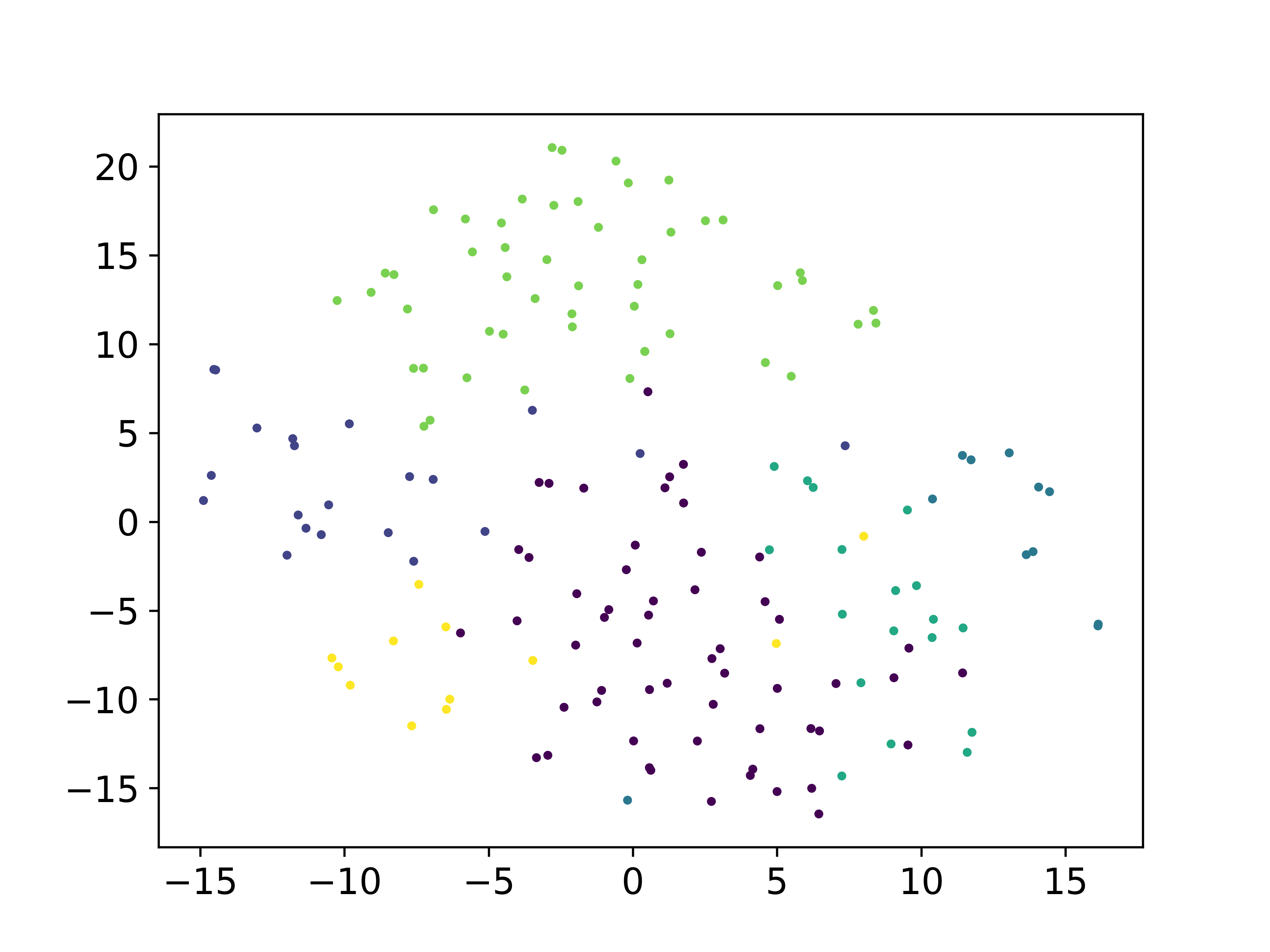}
         \caption{$\mathbf{X}$ of 3Sources}
     \end{subfigure}
     \begin{subfigure}[b]{0.23\textwidth}
         \centering
         \includegraphics[width=\linewidth]{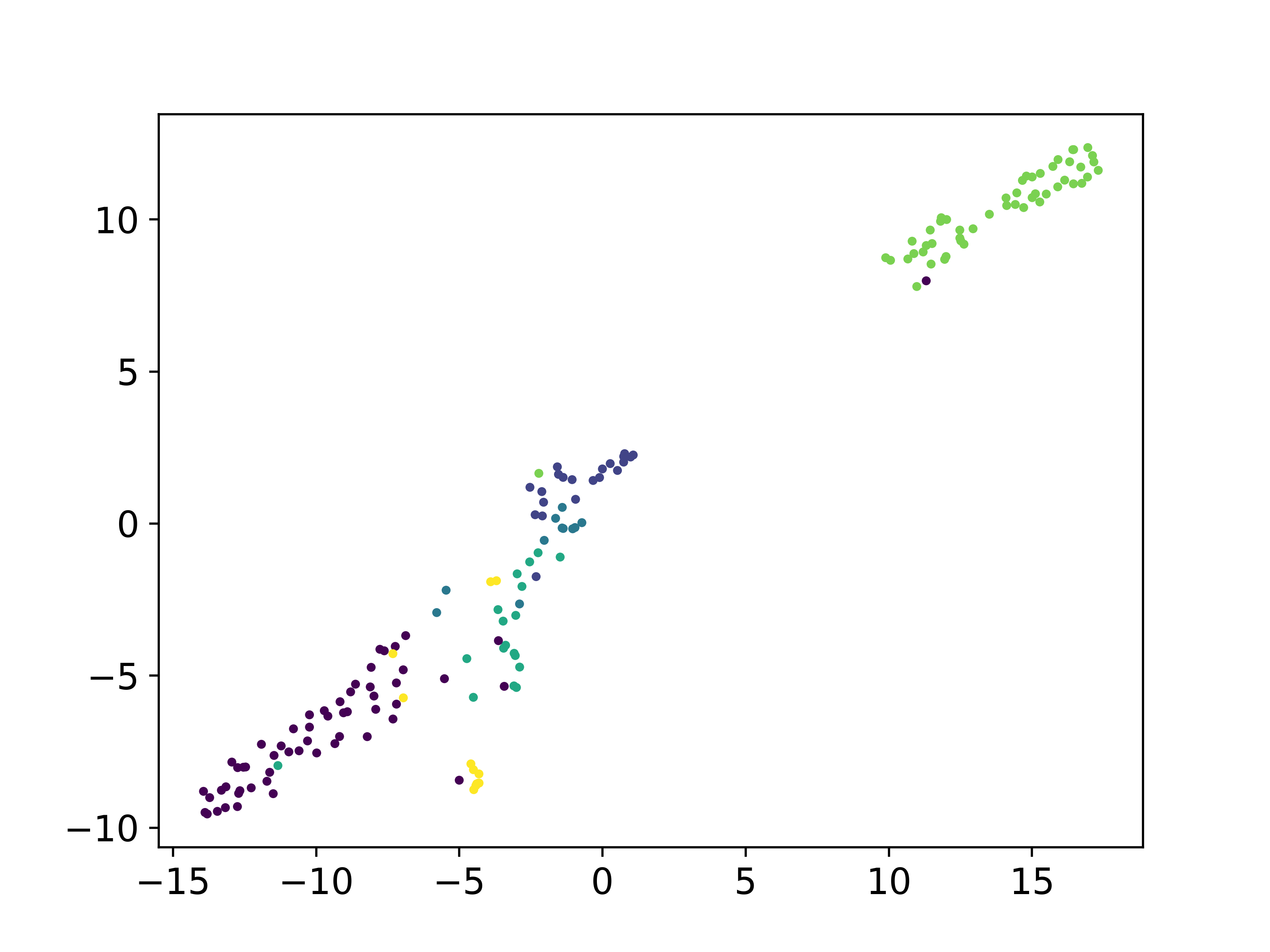}
         \caption{$\mathbf{Z}$ of 3Sources}
     \end{subfigure}
     \begin{subfigure}[b]{0.23\textwidth}
         \centering
         \includegraphics[width=\linewidth]{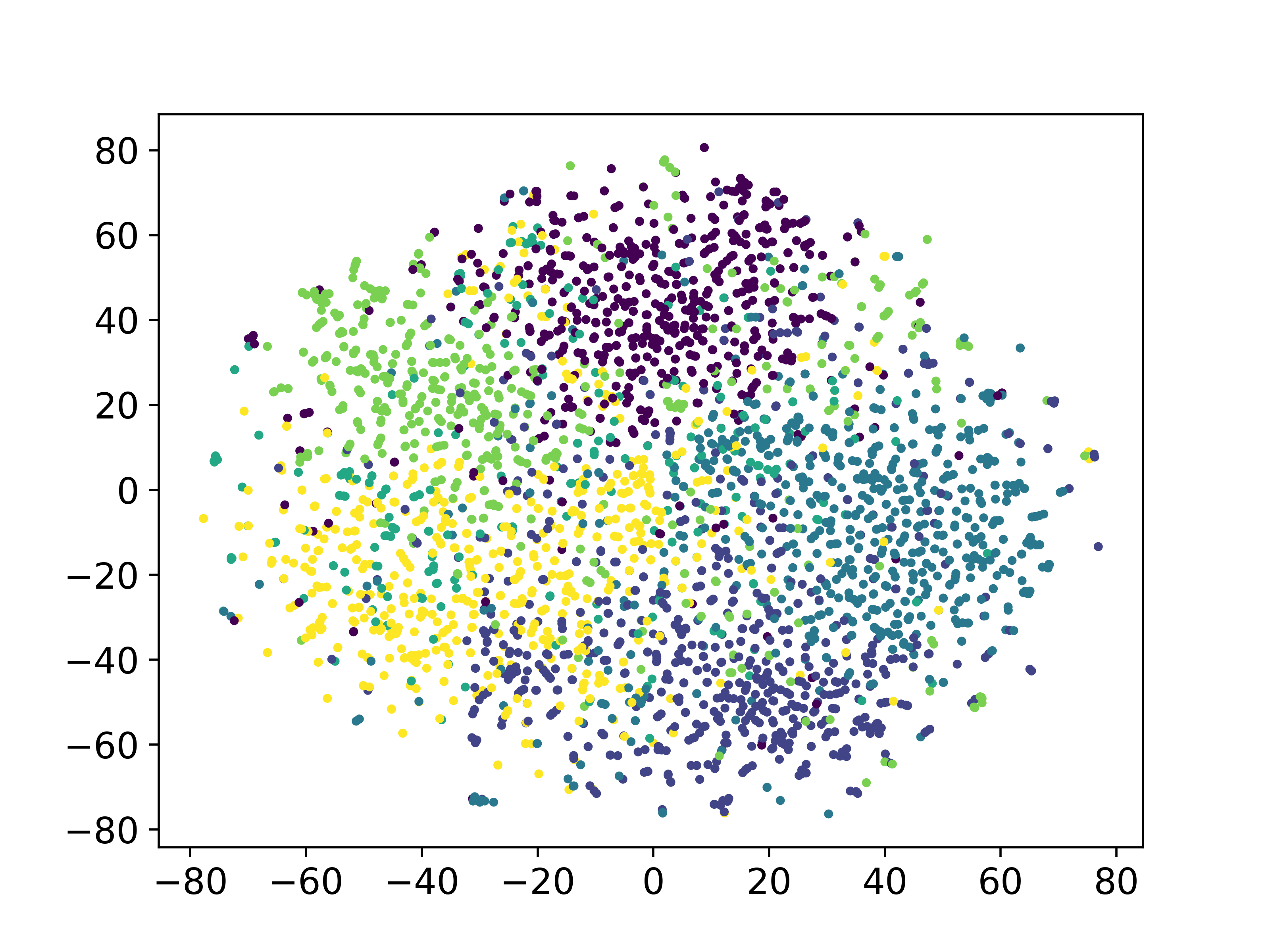}
         \caption{$\mathbf{X}$ of Citeseer}
     \end{subfigure}
     \begin{subfigure}[b]{0.23\textwidth}
         \centering
         \includegraphics[width=\linewidth]{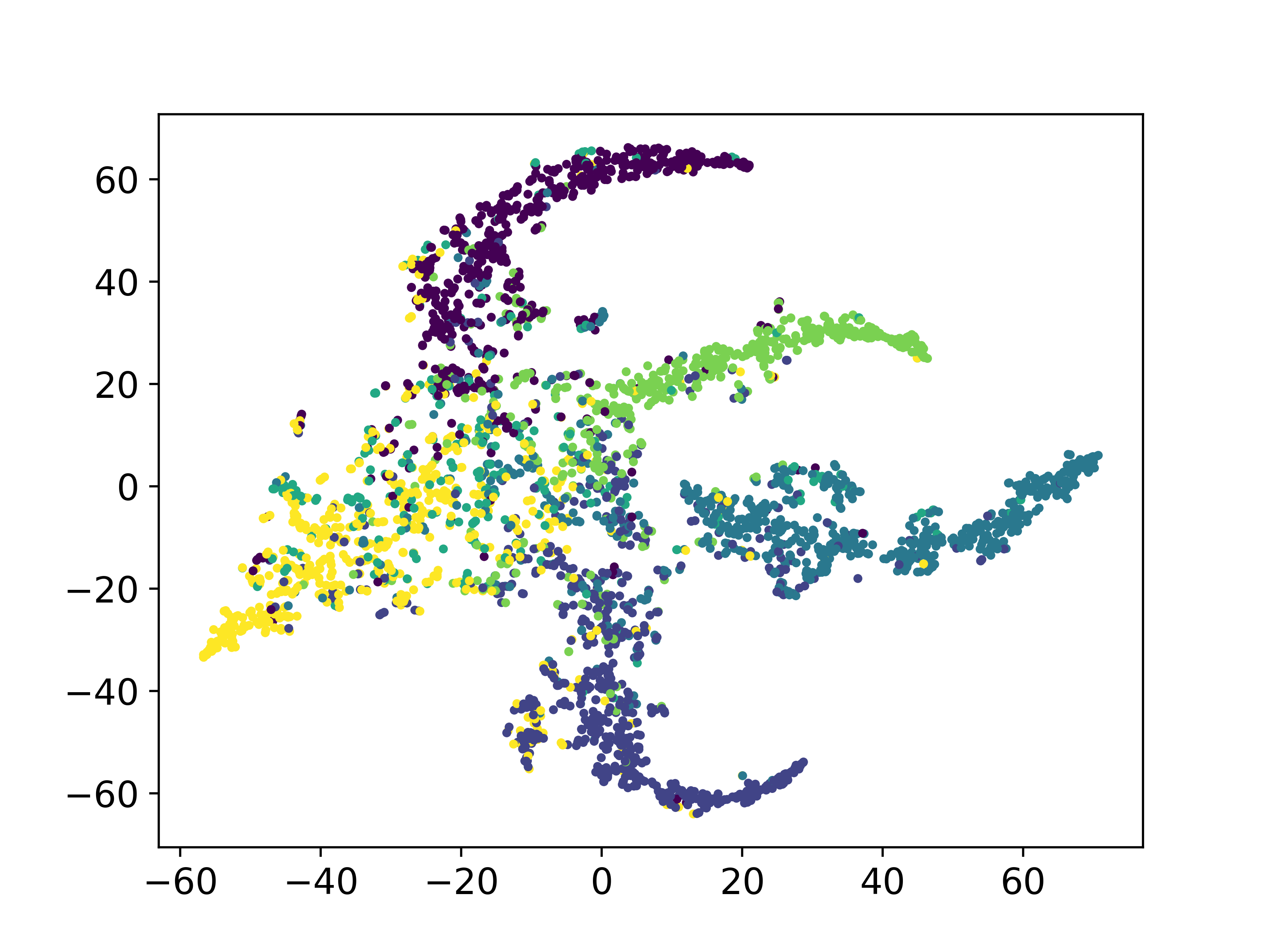}
         \caption{$\mathbf{Z}$ of Citeseer}
     \end{subfigure}
     \\
     \begin{subfigure}[b]{0.23\textwidth}
         \centering
         \includegraphics[width=\linewidth]{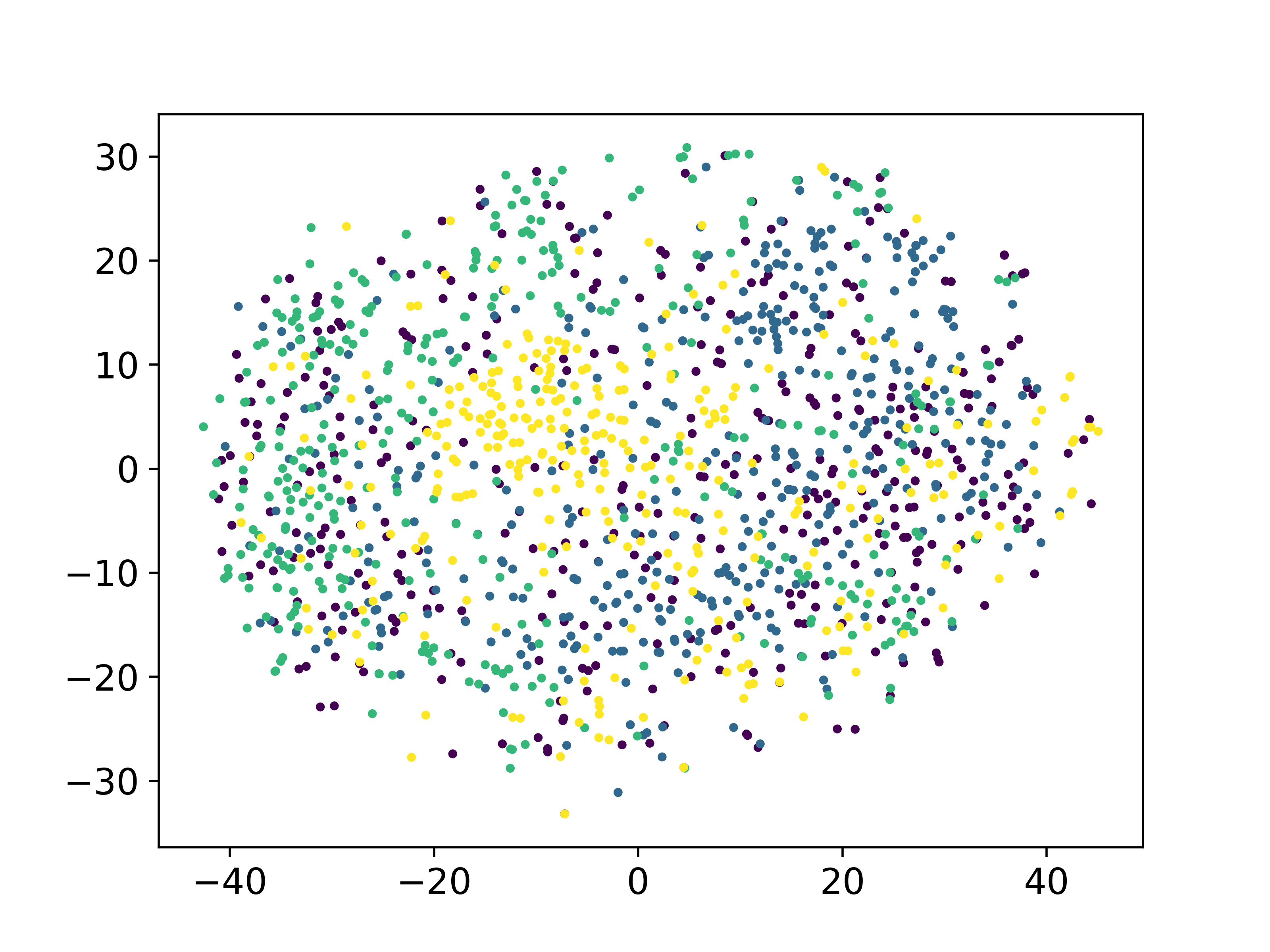}
         \caption{$\mathbf{X}$ of GRAZ02}
     \end{subfigure}
     \begin{subfigure}[b]{0.23\textwidth}
         \centering
         \includegraphics[width=\linewidth]{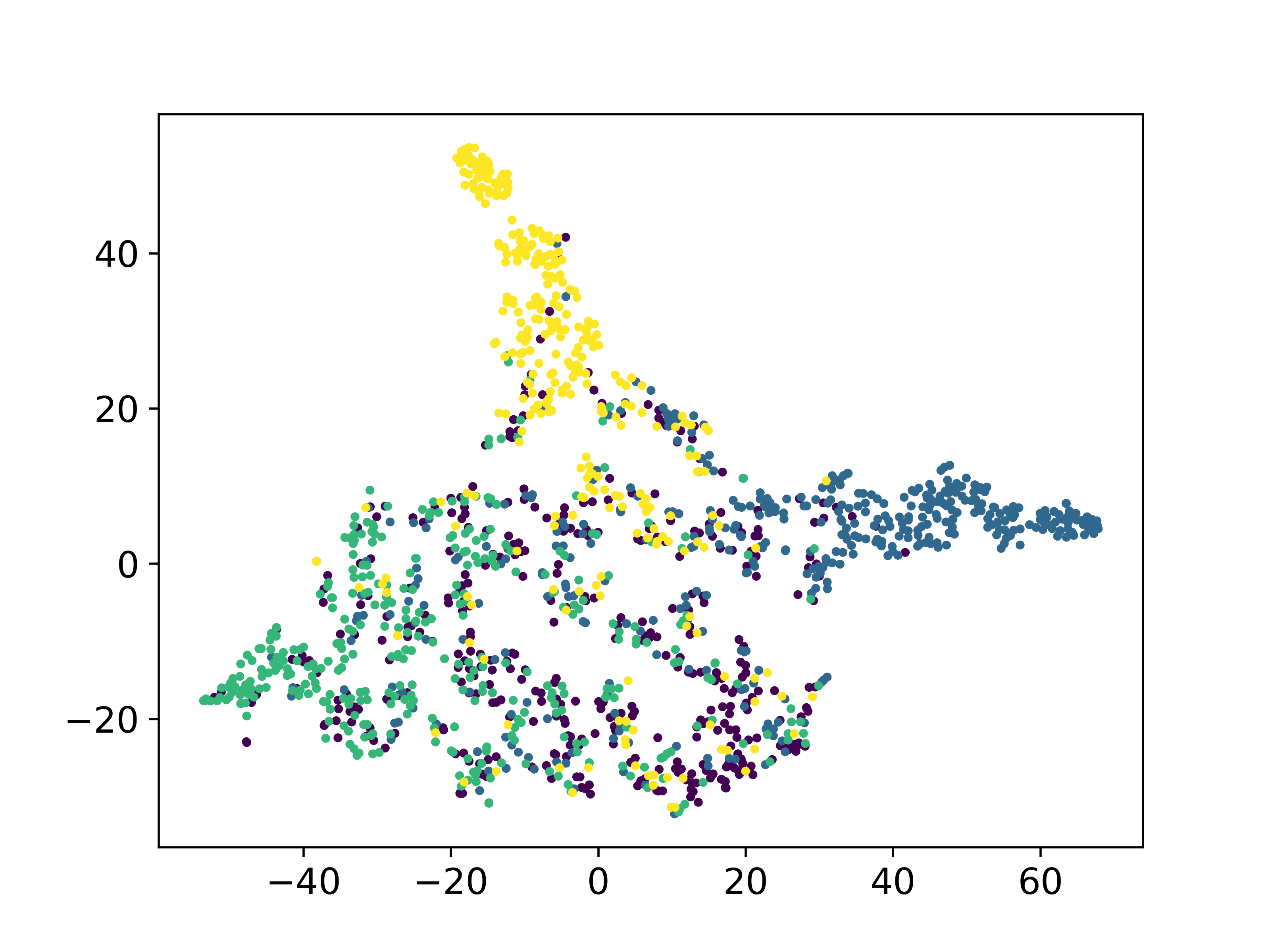}
         \caption{$\mathbf{Z}$ of GRAZ02}
     \end{subfigure}
     \begin{subfigure}[b]{0.23\textwidth}
         \centering
         \includegraphics[width=\linewidth]{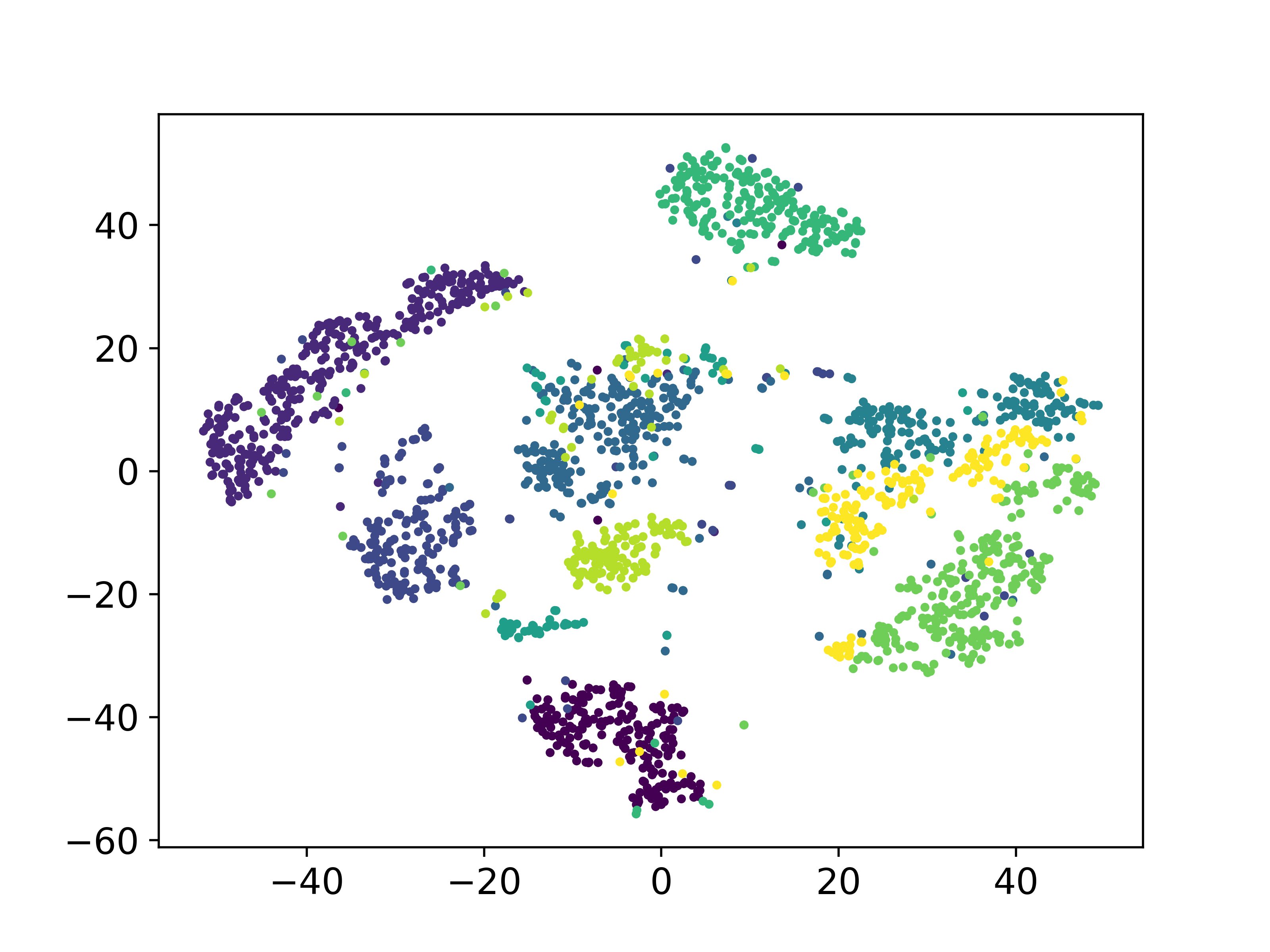}
         \caption{$\mathbf{X}$ of MNIST}
     \end{subfigure}
     \begin{subfigure}[b]{0.23\textwidth}
         \centering
         \includegraphics[width=\linewidth]{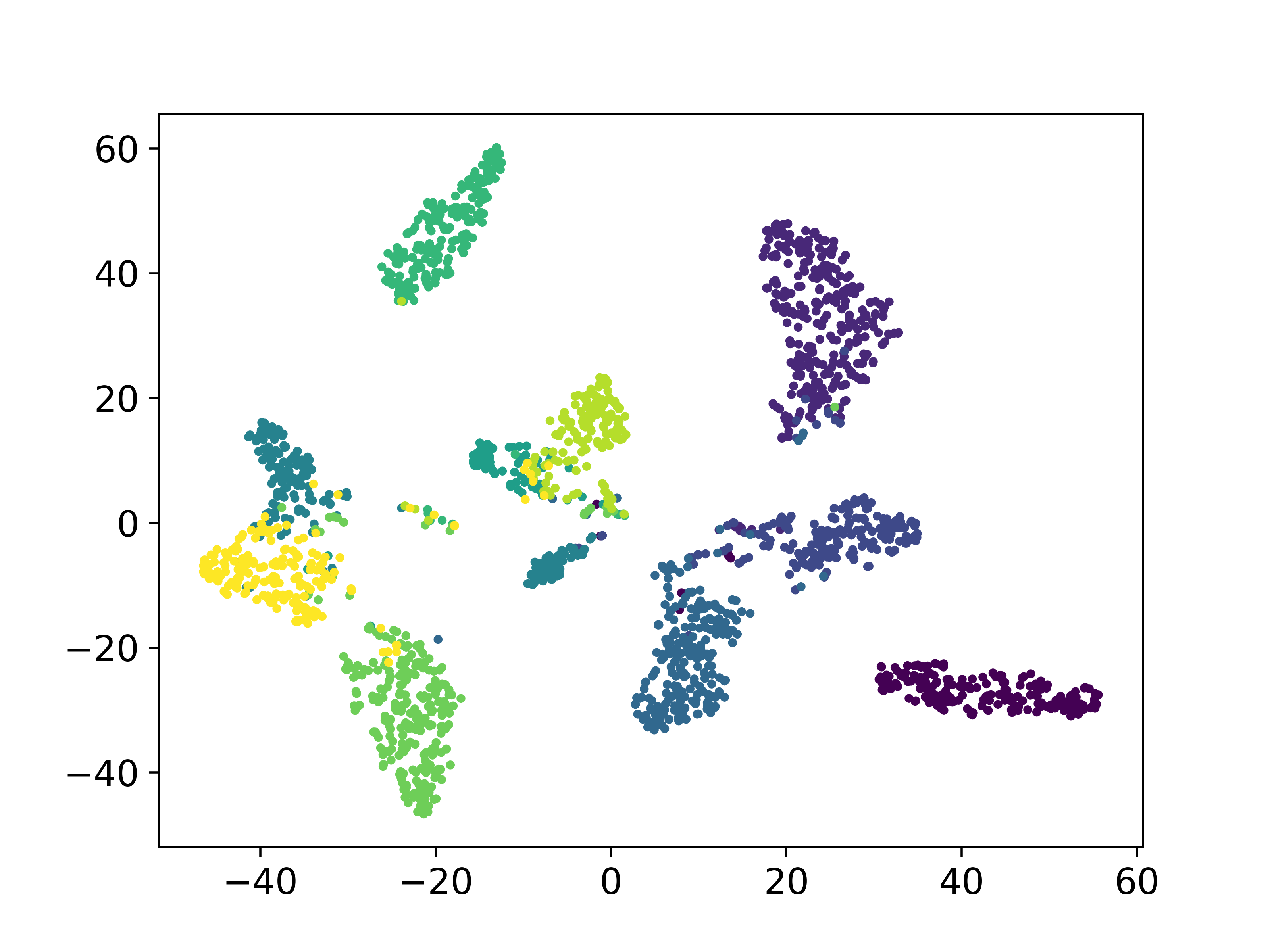}
         \caption{$\mathbf{Z}$ of MNIST}
     \end{subfigure}
     \\
     \begin{subfigure}[b]{0.23\textwidth}
         \centering
         \includegraphics[width=\linewidth]{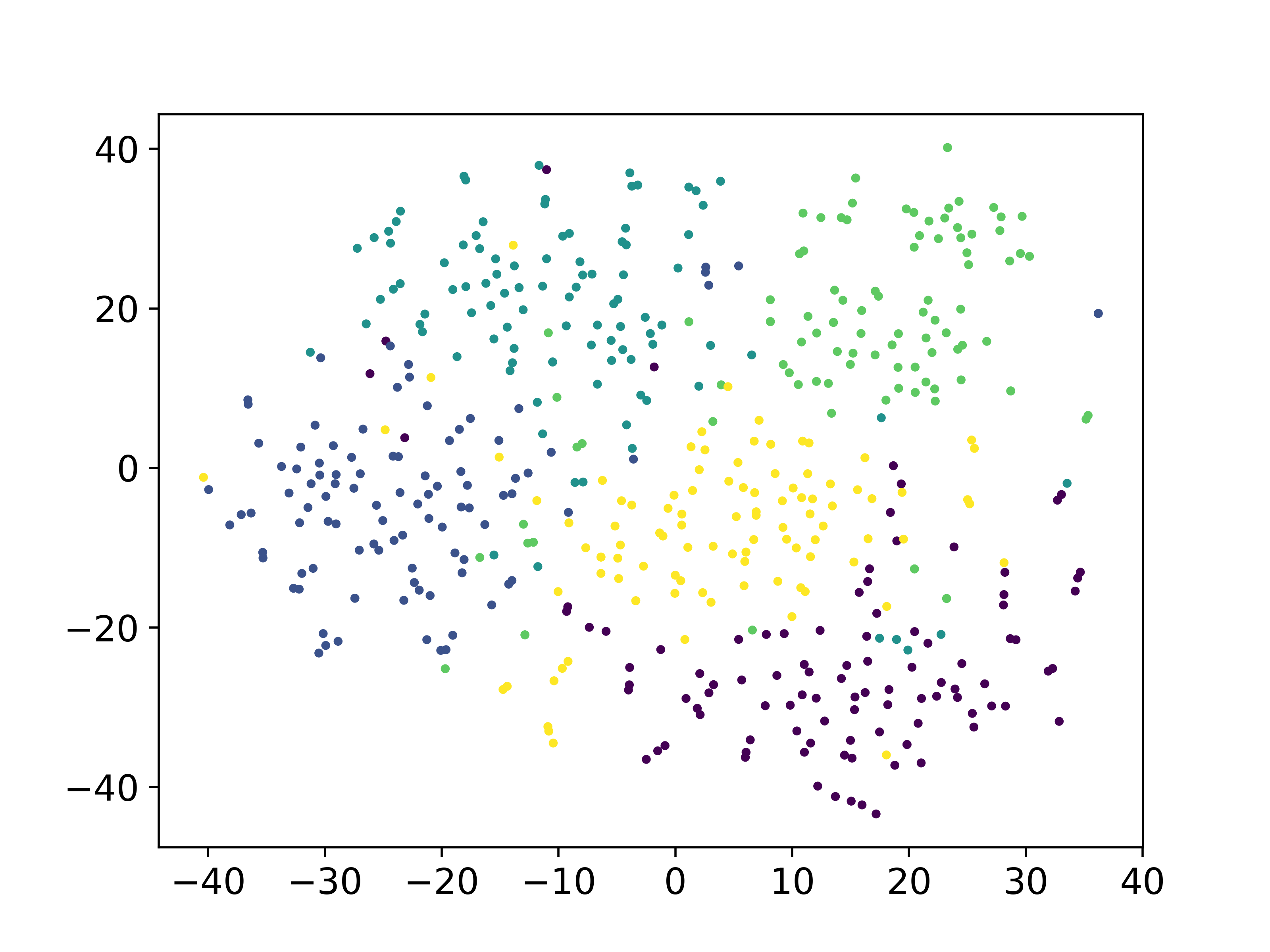}
         \caption{$\mathbf{X}$ of NGs}
     \end{subfigure}
     \begin{subfigure}[b]{0.23\textwidth}
         \centering
         \includegraphics[width=\linewidth]{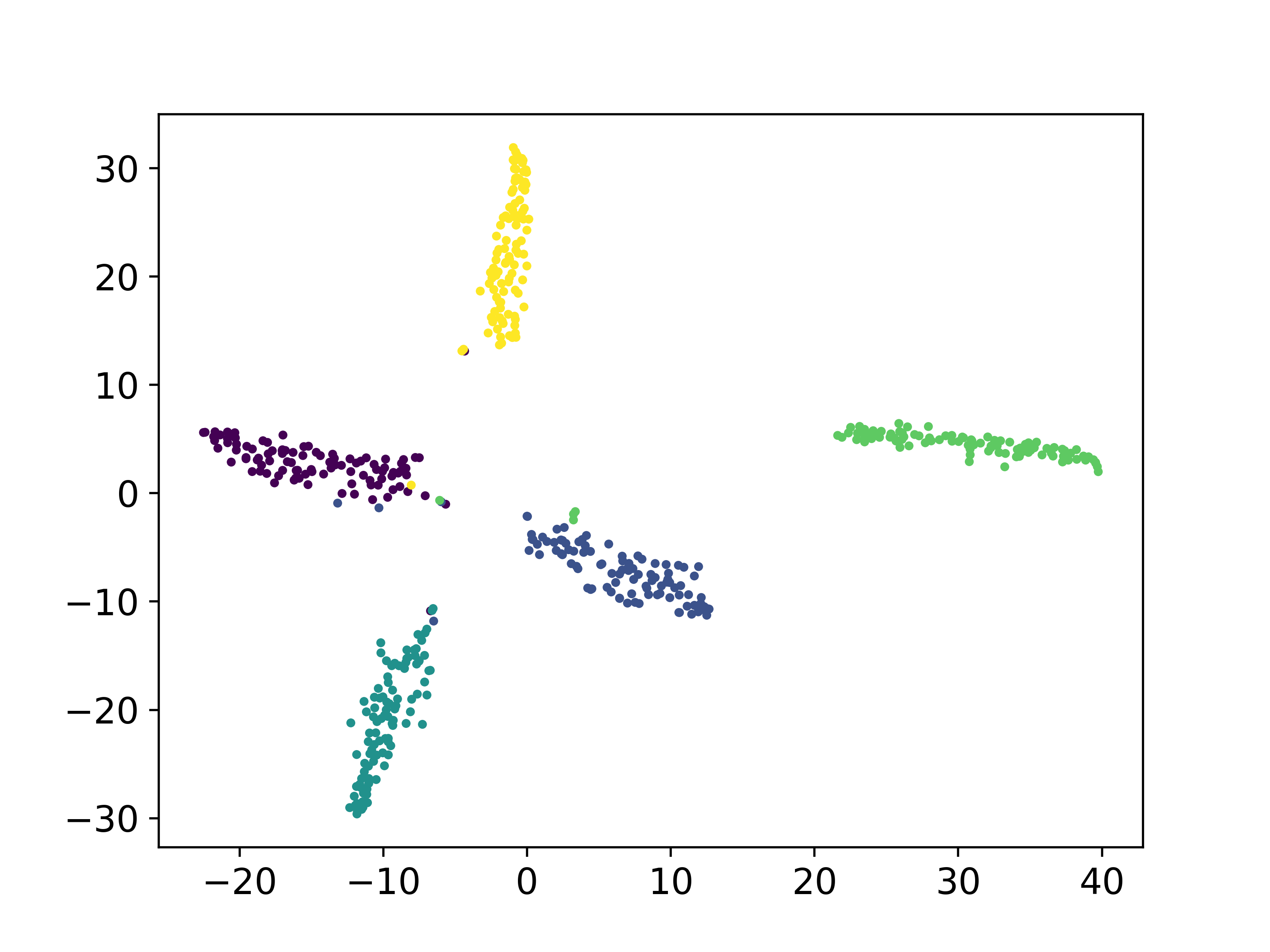}
         \caption{$\mathbf{Z}$ of NGs}
     \end{subfigure}
     \begin{subfigure}[b]{0.23\textwidth}
         \centering
         \includegraphics[width=\linewidth]{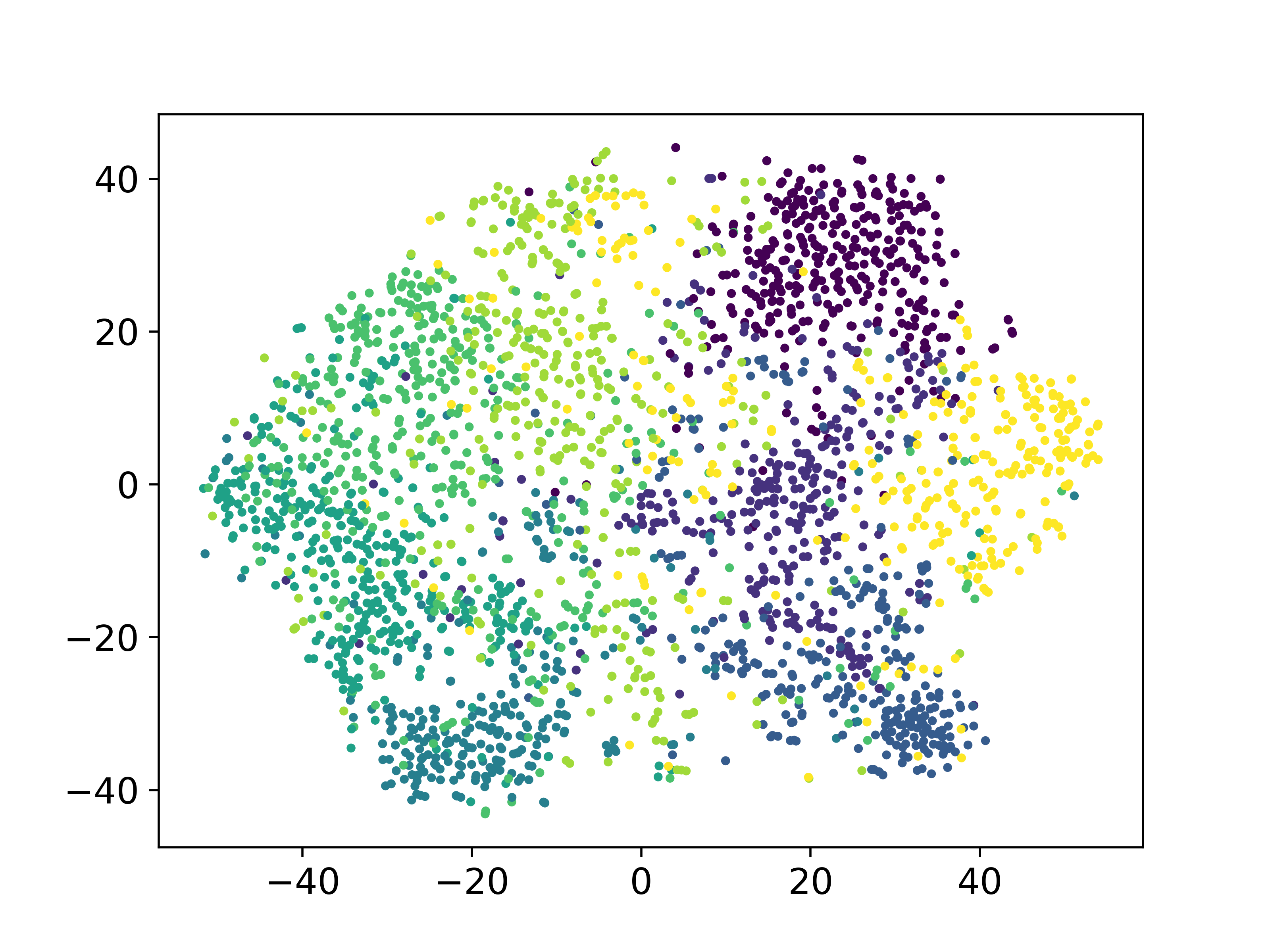}
         \caption{$\mathbf{X}$ of Out-Scene}
     \end{subfigure}
     \begin{subfigure}[b]{0.23\textwidth}
         \centering
         \includegraphics[width=\linewidth]{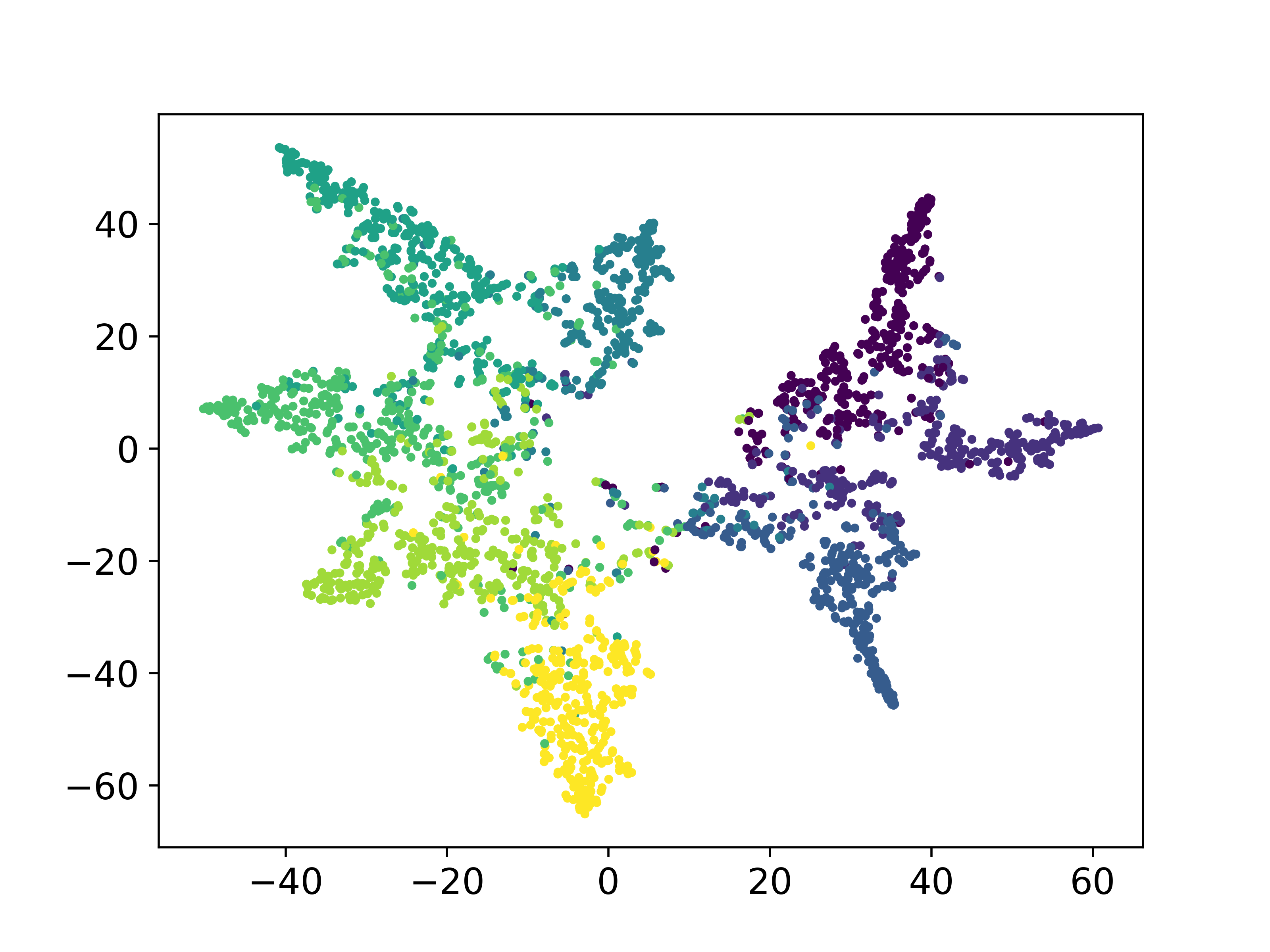}
         \caption{$\mathbf{Z}$ of Out-Scene}
     \end{subfigure}
        \caption{t-SNE projections of the input feature $\mathbf{X}$ and MV-SupGCN output $\mathbf{Z}$}
        \label{fig:t-sne}
\end{figure}

In the visualizations, each data point corresponds to a sample, with colors representing their true class labels. Ideally, samples from the same class should be grouped together in compact clusters, while samples from different classes should be distinctly separated. Fig. \ref{fig:t-sne} shows t-SNE visualizations of the original inputs and MV-SupGCN representations for six datasets.

As illustrated in Fig. \ref{fig:t-sne}, the contrastive learning framework employed by MV-SupGCN significantly improves the performance of semi-supervised classification. By comparing the distributions of the input features $\mathbf{X}$ and the learned representations $\mathbf{Z}$ on each dataset, MV-SupGCN clearly leads to more compact clusters within each class and better separation between different classes.

\normalsize\textit{\subsection{Convergence Analysis}}

\begin{figure}[h!]
    \centering
    \begin{subfigure}[b]{0.31\textwidth}  
        \centering
        \includegraphics[width=\linewidth]{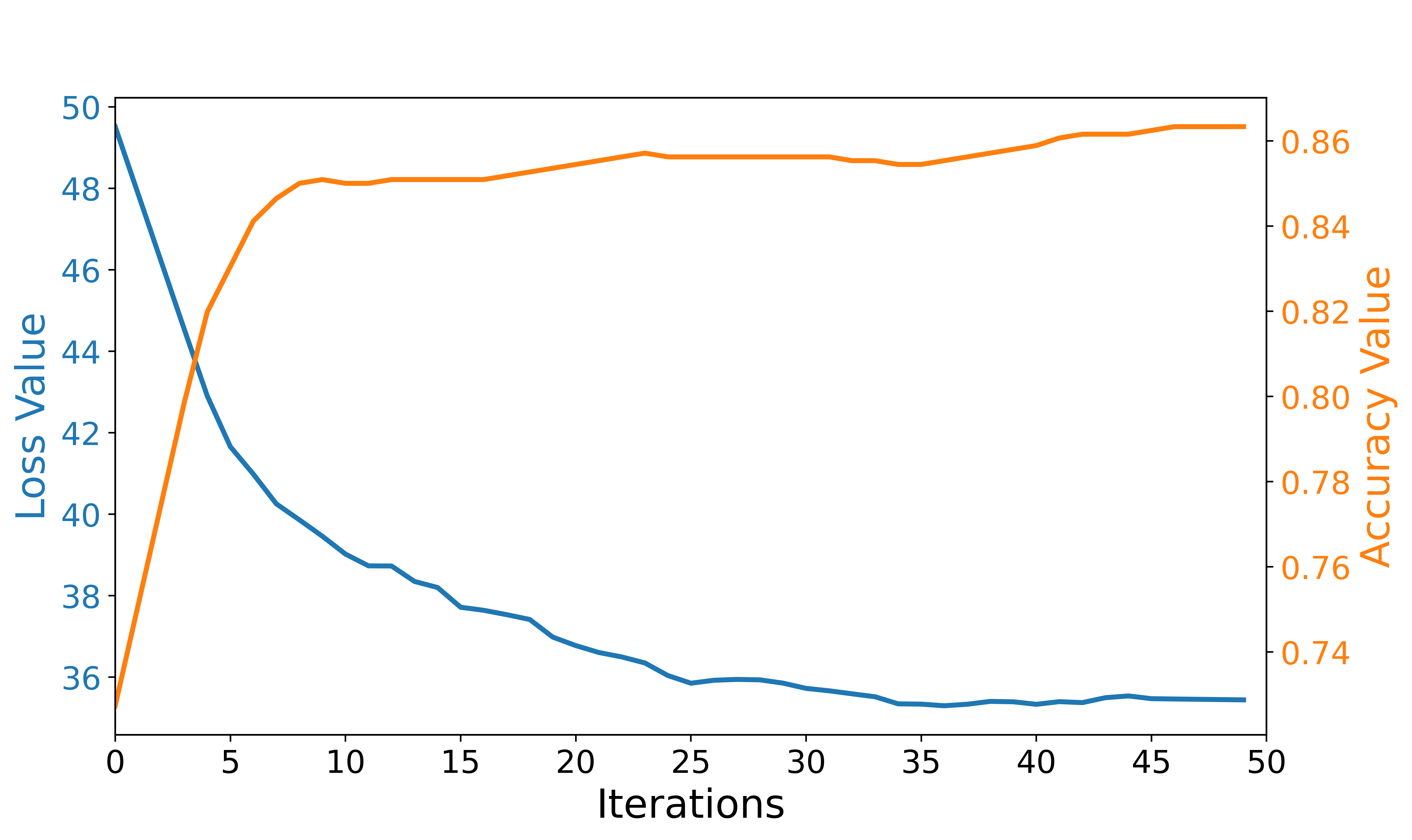}
        \caption{3Sources}
    \end{subfigure}
    \begin{subfigure}[b]{0.31\textwidth}
        \centering
        \includegraphics[width=\linewidth]{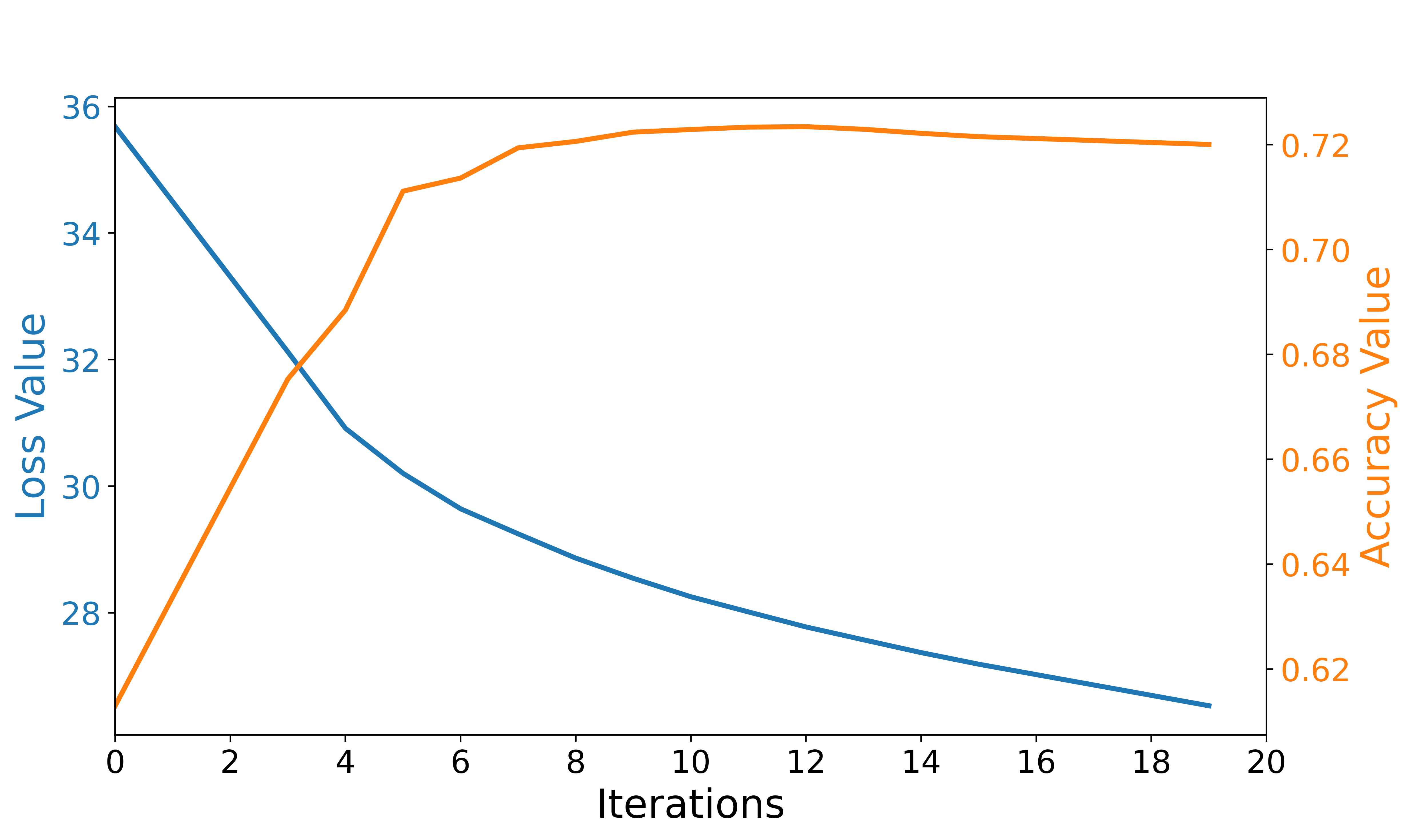}
        \caption{Citeseer}
    \end{subfigure}
    \begin{subfigure}[b]{0.31\textwidth}
        \centering
        \includegraphics[width=\linewidth]{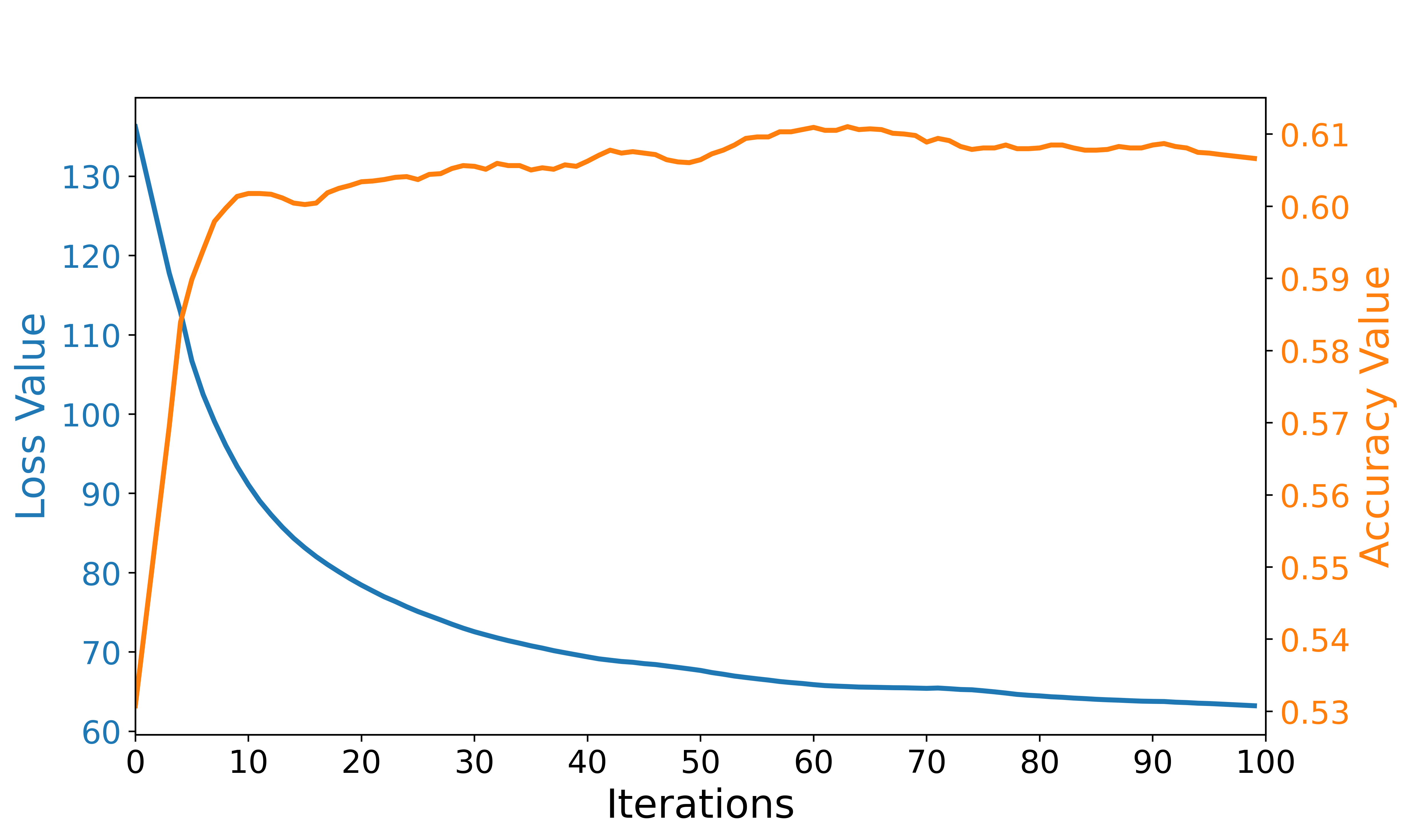}
        \caption{GRAZ02}
    \end{subfigure}
    
    \vskip\baselineskip  

    \begin{subfigure}[b]{0.31\textwidth}
        \centering
        \includegraphics[width=\linewidth]{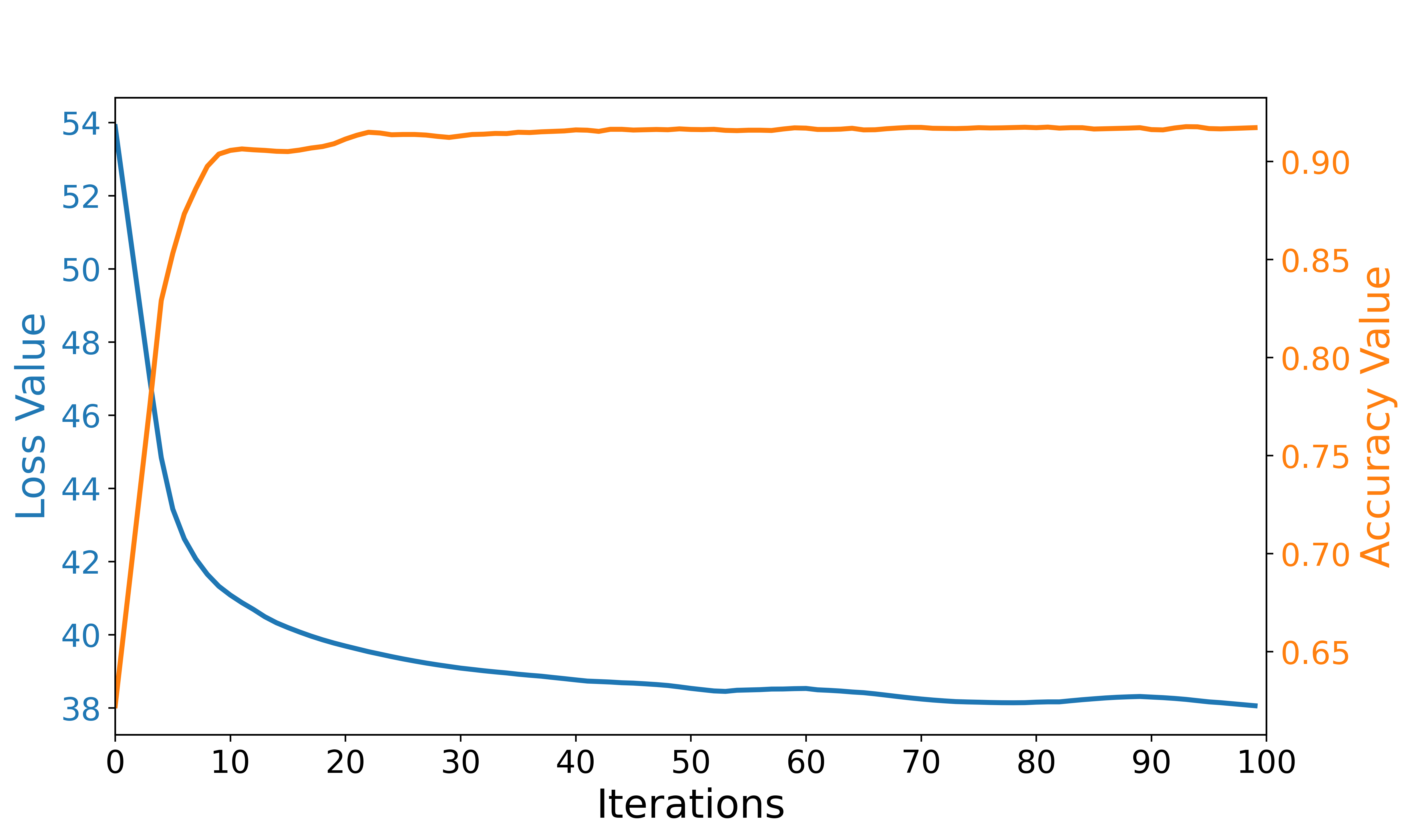}
        \caption{MNIST}
    \end{subfigure}
    \begin{subfigure}[b]{0.31\textwidth}
        \centering
        \includegraphics[width=\linewidth]{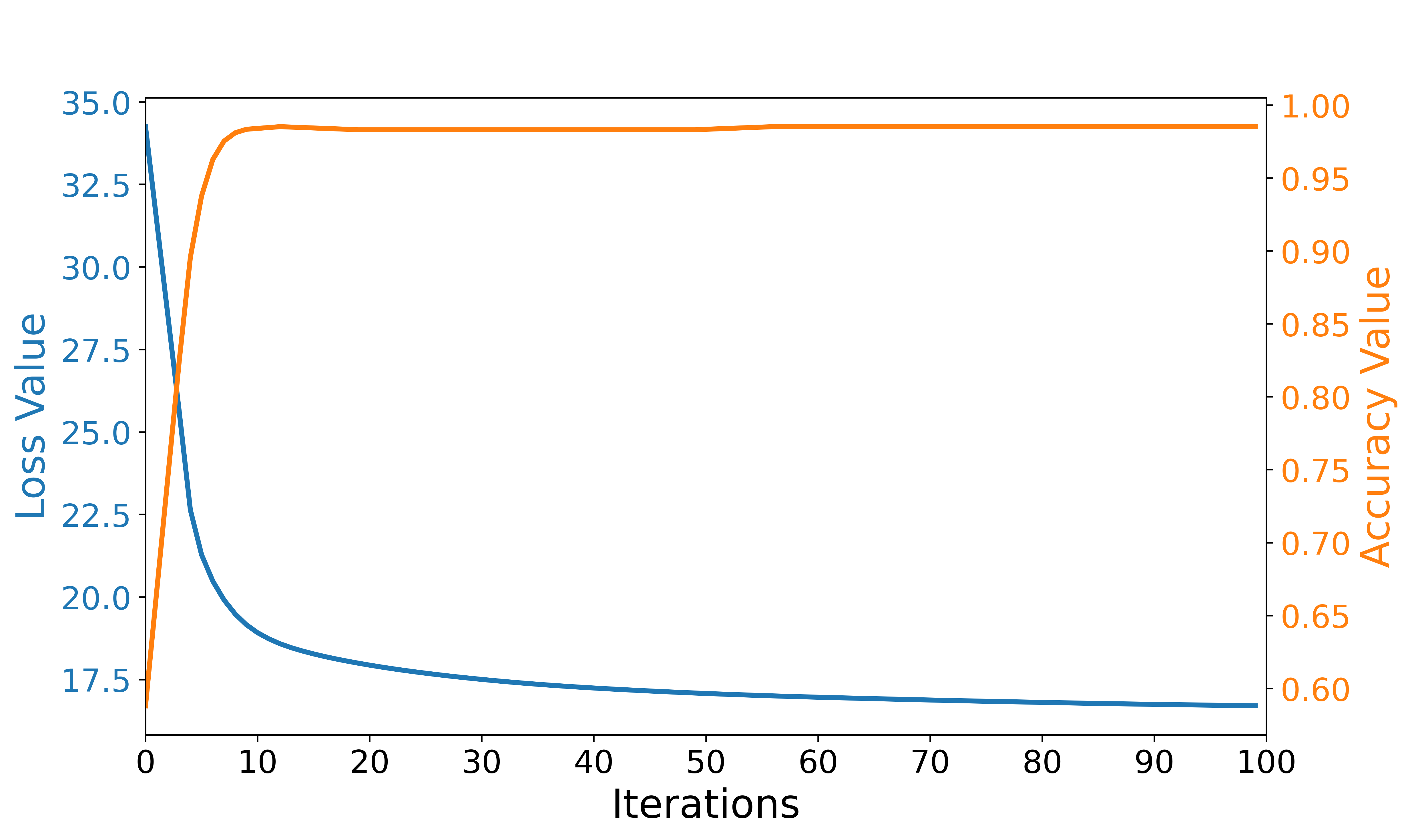}
        \caption{NGs}
    \end{subfigure}
    \begin{subfigure}[b]{0.31\textwidth}
        \centering
        \includegraphics[width=\linewidth]{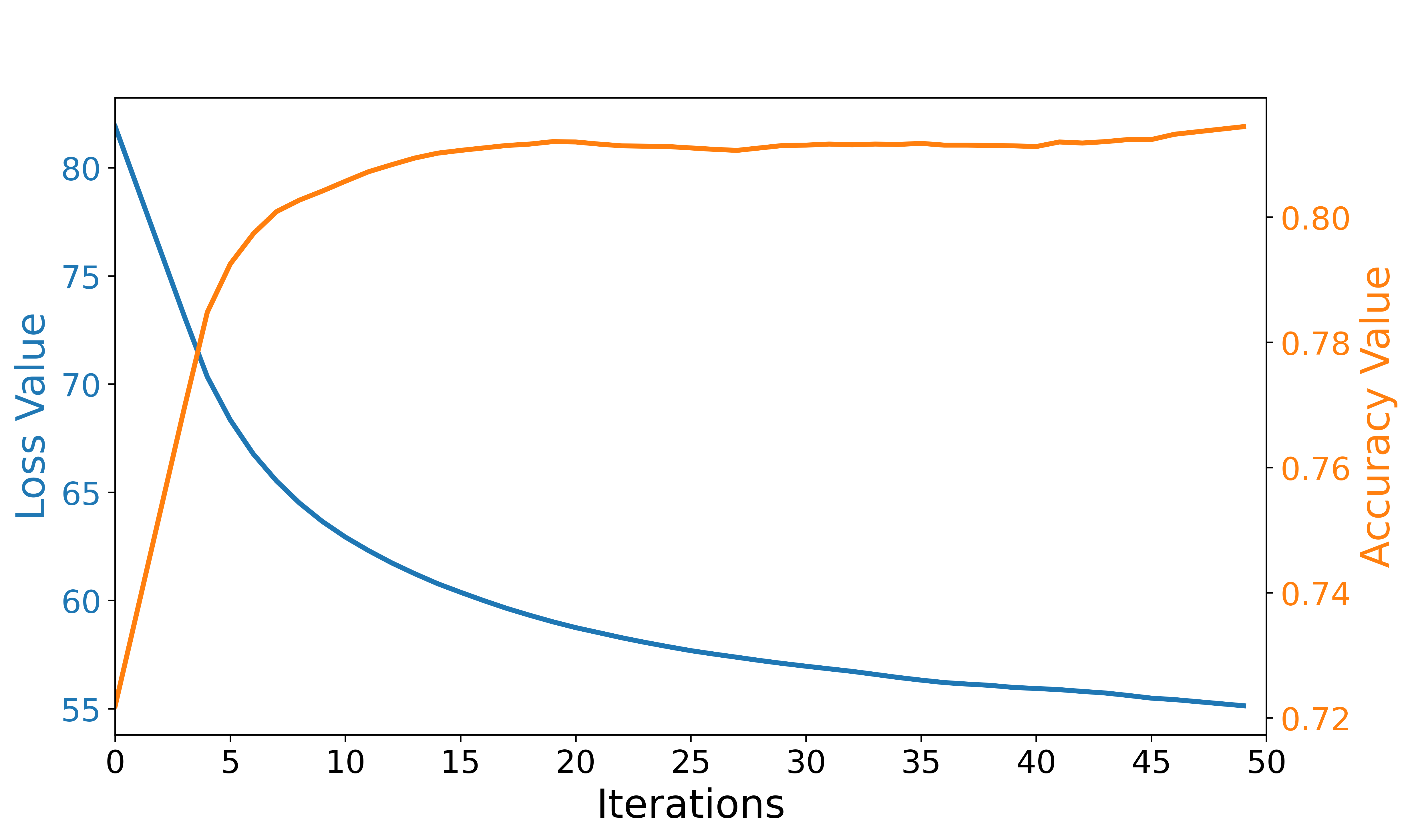}
        \caption{Out-Scene}
    \end{subfigure}

    \caption{Curves of loss value (black) and Accuracy value(orange) on six datasets.}
    \label{fig:convergence analysis}
\end{figure}

Fig. \ref{fig:convergence analysis} presents the evolution of accuracy and loss for MV-SupGCN during training on different datasets.

The MV-SupGCN model demonstrates strong convergence behavior by effectively optimizing multiple loss components simultaneously. The loss curve exhibits a rapid decrease during the early training epochs, eventually stabilizing as training progresses. Concurrently, the accuracy consistently improves in the initial stages and reaches a plateau, indicating convergence. These trends suggest that the model successfully learns meaningful feature representations and maintains robust performance under a combination of supervised and self-supervised learning signals.

\begin{figure}[h!]
\centering  

         \includegraphics[width=0.55\linewidth]{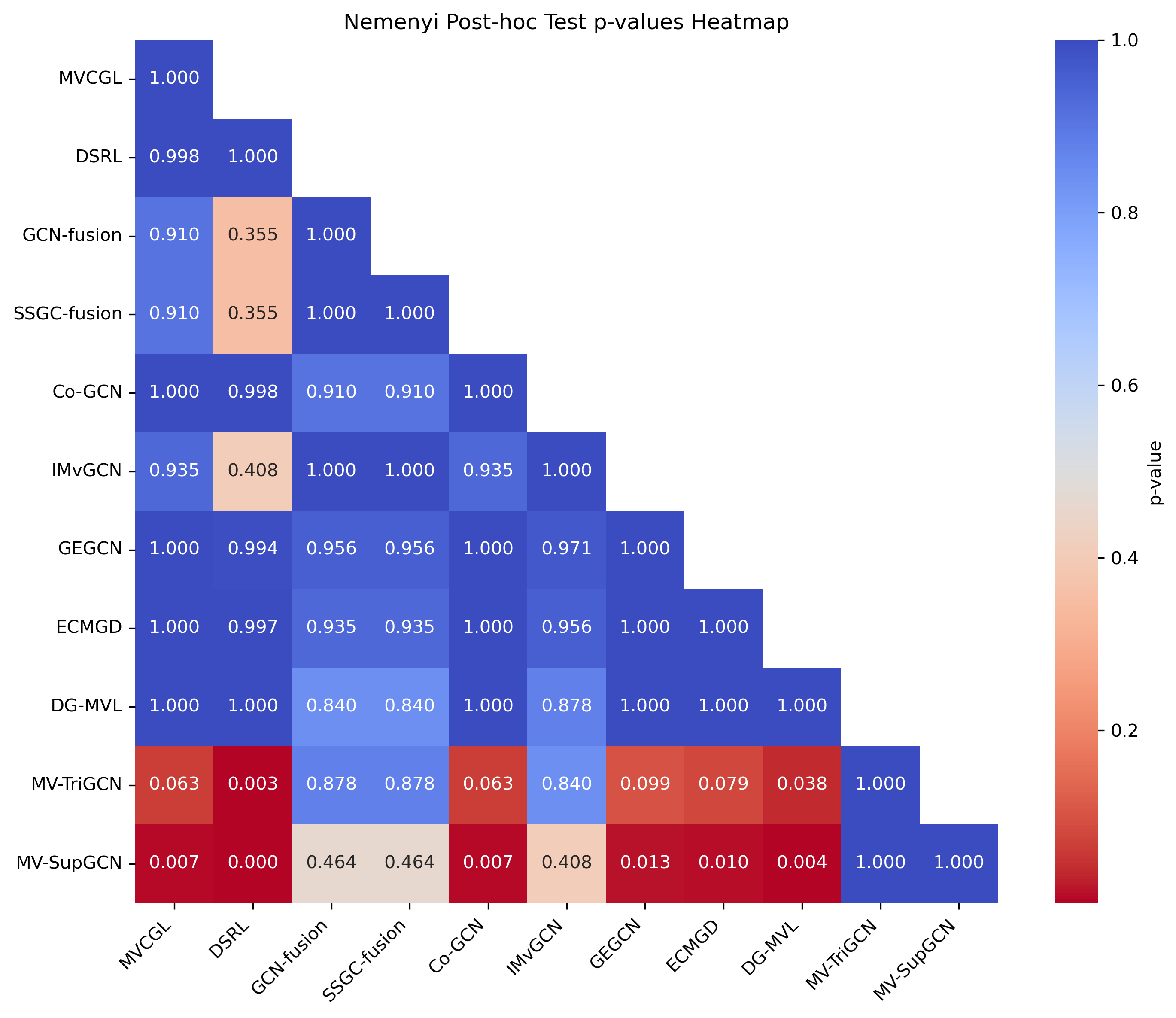}

    \caption{\textcolor{black}{Heatmap of pairwise $p$-values from the Nemenyi post-hoc test comparing multiple algorithms. }}
    \label{fig:heatmap}
\end{figure}

\begin{figure}[h]
\centering  

         \includegraphics[width=0.65 \linewidth]{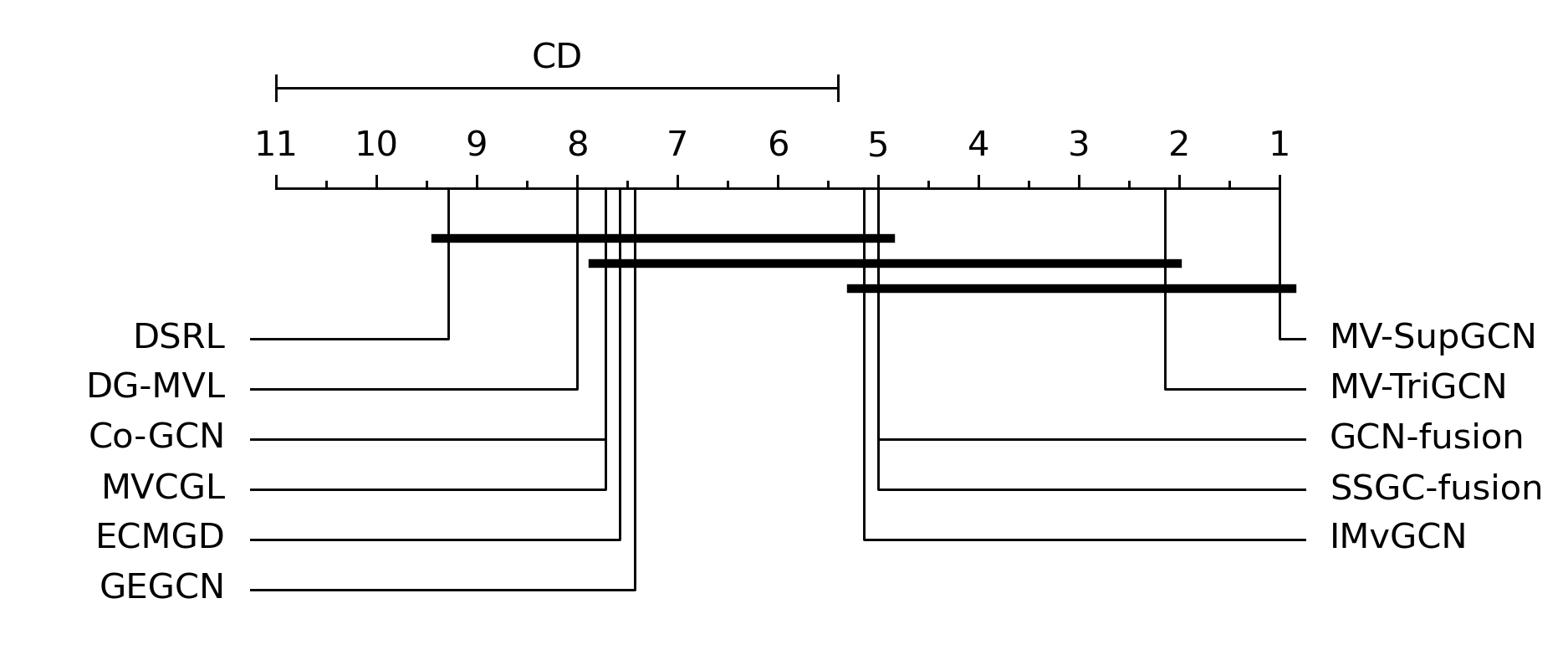}

    \caption{\textcolor{black}{Critical Difference diagram of the Nemenyi post-hoc test for 11 methods across seven datasets (CD = 5.60 at 0.05 significance level).}}
    \label{fig: CD}
\end{figure}

{\color{black}

\normalsize\textit{\subsection{Statistical Analysis}}

To rigorously evaluate the statistical significance of performance differences between our proposed MV-SupGCN framework and the baseline methods, we performed the Friedman test \cite{Friedman}. The method rankings used in the statistical analysis were determined based on the classification accuracies reported in Table \ref{tab:result}.

Specifically, given $K=11$ methods and $N=7$ datasets, the Friedman statistic is computed as follows:
\begin{equation}
\chi_{F}^{2} = \frac{12N}{K(K+1)} \left[ \sum_{j=1}^{K} R_{j}^{2} - \frac{K(K+1)^{2}}{4} \right]
\end{equation}
where $R_j$ denotes the average ranking of the $j$-th method across all datasets.

The null hypothesis of the Friedman test, which asserts that there are no significant differences in the central tendencies among the compared methods, is rejected with a $p$-value less than $10^{-5}$, indicating a statistically significant difference in performance across the evaluated methods.

Additionally, we use the post-hoc Nemenyi test \cite{nemenyi} to determine which differences are statistically significant. Fig. \ref{fig:heatmap} presents a heatmap of pairwise p-values from the Nemenyi post-hoc test comparing multiple algorithms. Each cell indicates the significance level of the difference between two algorithms, with color intensity representing the magnitude of the $p$-values (cool colors correspond to higher $p$-values, i.e., less significant differences, while warm colors correspond to lower p-values, i.e., more significant differences). It can be observed that, except for MV-TriGCN, our method exhibits significant differences compared to all other methods.

Finally, Fig. \ref{fig: CD} depicts the Critical Difference (CD) diagram from the Nemenyi post-hoc test conducted on 11 methods across seven datasets (with a CD value of 5.60 at the 0.05 significance level). The results indicate that our proposed MV-TriGCN method attains the highest average ranking among all evaluated approaches. While the statistical significance analysis shows that performance differences among the top-ranking methods are generally subtle and may not always reach strict significance thresholds, the corresponding heatmap in Fig. \ref{fig:heatmap} reveals that MV-TriGCN exhibits notably significant differences when compared to most other methods. Overall, MV-TriGCN consistently demonstrates robust and sustained superior performance across all tested datasets and encoder architectures, achieving the best overall average ranking. We emphasize that the computed Critical Difference (CD) is relatively large since the number of tested datasets is small. }

\normalsize\textit{\subsection{Ablation Study}}

In this section, we conducted an ablation study to evaluate the individual contributions of the four key components in our proposed method. Specifically, we examined the fusion of $2V$ graphs, denoted as \textbf{G} (comprising $V$ KNN graphs and $V$ semi-supervised graphs), the Supervised Contrastive loss (\textbf{$\mathbf{L}_{SupCon}$}), the Self-Supervised Contrastive loss (\textbf{$\mathbf{L}_{SelfCon}$}), as well as the use of pseudo-labels to enhance the performance of CE loss, SupCon loss, and SelfCon loss.

The results presented in Table \ref{tab:ablation} indicate that fusing semi-supervised graphs with KNN graphs (i.e., utilizing $2V$ graphs) enhances model performance. Moreover, incorporating the proposed SupCon loss further improves accuracy. The combination of KNN and semi-supervised graphs, together with SupCon loss, SelfCon loss, and the use of pseudo-labels, each contribute to performance gains. Overall, all components positively impact the framework, and their joint application achieves the best accuracy in semi-supervised tasks.


\begin{table}[htbp]
\centering
\caption{Ablation Results for the MV-SupGCN Method.}
\renewcommand{\arraystretch}{1.5}
\resizebox{0.75\textwidth}{!}
{
\begin{tabular}{ccccll} 
\hline
$\mathbf{2 graphs/view}$ &$\mathbf{L_{SupCon}}$&$\mathbf{L_{SelfCom}}$&$\mathbf{pseudo}$&Citeseer &MNIST  \\ 
\hline
$\times$ &$\times$ &$\times$&$\times$ &68.13(0.85) &89.56 (0.65)  \\
$\checkmark$ &$\times$ &$\times$&$\times$ &69.49(0.82) &90.11(0.36)  \\
\hline
$\checkmark$ &$\checkmark$ &$\times$&$\times$ &70.14(0.98) &91.07(0.70)  \\
$\checkmark$ &$\times$ &$\checkmark$&$\times$&70.79(0.62) &90.18(0.34)  \\
$\checkmark$ &$\checkmark$ &$\checkmark$&$\times$&70.84(0.71) &91.16 (0.54)  \\
$\checkmark$ &$\times$ &$\times$&$\checkmark$ &70.48(0.85) &90.54(0.53) \\
$\checkmark$ &$\checkmark$ &$\times$&$\checkmark$&71.04(0.91) & 91.15(0.89)  \\
$\checkmark$ &$\times$ &$\checkmark$&$\checkmark$ &71.24(0.70)&90.82(0.23)  \\
$\checkmark$ &$\checkmark$ &$\checkmark$&$\checkmark$&\textbf{71.50(0.82)} &\textbf{91.38(0.56)}
  \\
 \hline 
\end{tabular}
}
\label{tab:ablation} 
\end{table}

\section{Conclusion}
\label{sec:conclusion}

The proposed Mv-SupGCN framework demonstrates significant improvements in model generalization and performance by leveraging both labeled and unlabeled data. By incorporating a joint loss function combining Cross-Entropy and SupCon losses, Mv-SupGCN effectively measures intra- and inter-class distances, enhancing feature learning. The use of two distinct graph construction methods, KNN and semi-supervised graph, provides a more stable and diverse hypothesis space, reducing generalization error. Moreover, the framework simultaneously applies contrastive learning and pseudo-labeling to unlabeled samples, fully harnessing unlabeled data to boost the model’s effectiveness. Experimental results on multiple datasets show that Mv-SupGCN outperforms existing graph-based algorithms, highlighting its superior accuracy, stability, and generalization ability.

Experimental results across multiple datasets demonstrate that Mv-SupGCN surpasses current graph-based algorithms, showcasing its superior accuracy, stability, and generalization capabilities, thus emphasizing the method's practicality in real-world applications. For example, the fusion of Lidar point clouds and RGB images in a multi-view scene enhances autonomous vehicles' perception by combining precise depth information with detailed semantic context, improving both performance and safety.

\textcolor{black}{As future work, we plan to explore refined semi-supervised strategies like entropy minimization and consistency regularization within the multi-view contrastive learning framework to enhance generalization and semantic alignment. We also aim to investigate incorporating warm-up schedules to gradually introduce pseudo-labeled data, improving the model’s adaptability and robustness to pseudo-label quality variations.}  We envision evaluating our method with different GNN backbones, such as GAT and GraphSAGE. This can offer insights into its generalizability and further demonstrate the flexibility and applicability of our framework across diverse graph learning tasks.



\end{document}